\documentclass[]{interact}

\usepackage{epstopdf}
\usepackage{subcaption}

\usepackage{natbib}
\bibpunct[, ]{(}{)}{;}{a}{}{,}
\renewcommand\bibfont{\fontsize{10}{12}\selectfont}

\theoremstyle{plain}

\theoremstyle{definition}

\theoremstyle{remark}

\usepackage{lineno,hyperref}
\modulolinenumbers[5]

\usepackage{url}
\usepackage{graphicx}
\usepackage{booktabs}

\usepackage{tabularx}
\usepackage{amssymb}
\usepackage{graphicx}
\usepackage{xcolor,colortbl}

\usepackage{multirow}
\usepackage{mathtools}
\usepackage{siunitx}
\usepackage{url}

\usepackage{algorithmic}

\DeclarePairedDelimiter\floor{\lfloor}{\rfloor}
\newcolumntype{Z}{>{\raggedright\let\newline\\\arraybackslash\hspace{0pt}}X}

\begin{document}

\articletype{ARTICLE TEMPLATE}

\title{Visual descriptors for content-based retrieval of remote sensing images}

\author{
\name{Paolo Napoletano~\textsuperscript{a}\thanks{CONTACT P. Napoletano. Email: paolo.napoletano@disco.unimib.it}}
\affil{\textsuperscript{a}Department of Informatics, Systems and Communication. University of Milano - Bicocca. Viale Sarca 336/U14. 20126 Milan, Italy}
}

\maketitle

\begin{abstract}
In this paper we present an extensive evaluation of visual descriptors for the content-based retrieval of remote sensing (RS) images. The evaluation includes global hand-crafted, local hand-crafted, and Convolutional Neural Network (CNNs) features coupled with four different Content-Based Image Retrieval schemes. We conducted all the experiments on two publicly available datasets: the 21-class UC Merced Land Use/Land Cover (LandUse) dataset and 19-class High-resolution Satellite Scene dataset (SceneSat). The content of RS images might be quite heterogeneous, ranging from images containing fine grained textures, to coarse grained ones or to images containing objects. It is therefore not obvious in this domain, which descriptor should be employed to describe images having such a variability.  Results demonstrate that CNN-based features perform better than both global and and local hand-crafted features whatever is the retrieval scheme adopted. Features extracted from SatResNet-50, a residual CNN suitable fine-tuned on the RS domain, shows much better performance than a residual CNN pre-trained on multimedia scene and object images. Features extracted from NetVLAD, a CNN that considers both CNN and local features, works better than others CNN solutions on those images that contain fine-grained textures and objects.
\end{abstract}


\section{Introduction}
The recent availability of a large amount of remote sensing (RS) images is boosting the design of systems for their management. A conventional RS image management system usually exploits  \emph{high-level} features to index the images such as textual annotations and metadata~\cite{datta2008image}. 
In the recent years, researchers are focusing their attention on systems that exploit  \emph{low-level} features extracted from images for their automatic indexing and retrieval ~\cite{Jain1998}. These types of systems are known as Content-Based Image Retrieval (CBIR) systems and they have demonstrated to be very useful in the RS domain~\cite{Bruzzone,aptoula2014,ozkan2014,yang2013,zajic2007accelerating}. 

The CBIR systems allow to search and retrieve images that are similar to a given query image~\cite{smeulders2000content,datta2008image}. Usually their performance strongly depends on the effectiveness of the features exploited for representing the visual content of the images~\cite{smeulders2000content}. The content of RS images might be quite heterogeneous, ranging from images containing fine grained textures, to coarse grained ones or to images containing objects~\cite{yang2010bag,dai2011satellite}. It is therefore not obvious in this domain, which descriptor should be employed to describe images having such a variability.

In this paper we compare several visual descriptors in combination with four different retrieval schemes. Such descriptors can be grouped in two classes. The first class includes traditional global hand-crafted descriptors that were originally designed for image analysis and local hand-crafted features that were originally designed for object recognition. The second class includes features that correspond to intermediate representations of Convolutional Neural Networks (CNNs) trained for generic object and/or scene and RS image recognition.

To reduce the influence of the retrieval scheme on the evaluation of the features we investigated the features coupled with four different image retrieval schemes. The first one, that is also the simplest one, is a \emph{basic} image retrieval system  that takes one image as input query and returns a list of images ordered by their degree of feature similarity. The second and the third ones, named \emph{pseudo} and \emph{manual Relevance Feedback} (RF), extend the \emph{basic} approach by expanding the initial query. The \emph{Pseudo RF} scheme uses the $n$ most similar images to the initial query, for re-querying the image database. The final result is obtained by combining the results of each single query. In the \emph{manual RF}, the set of relevant images is suggested by the user which evaluates the result of the initial query. The last scheme considered is named \emph{active-learning-based RF}~\cite{Bruzzone}. It exploits Support Vector Machines (SVM) to classify relevant and not relevant images on the basis of the user feedback. 

For the sake of completeness, for the first three retrieval schemes we considered different measure of similarity, such as Euclidean, Cosine, Manhattan, and $\chi^2$-square, while for the \emph{active-learning-based RF} scheme we considered the histogram intersection as similarity measure, as proposed by the original authors~\cite{Bruzzone}. 

We conducted all the experiments on two publicly available datasets: the 21-class UC Merced Land Use/Land Cover dataset~\cite{yang2010bag} (LandUse) and 19-class High-resolution Satellite Scene dataset~\cite{dai2011satellite} (SatScene). Evaluations exploit several computational measures in order to quantify the effectiveness of the features. To make the experiments replicable, we made publicly available all the visual descriptors calculated as well as the scripts for making the evaluation of all the image retrieval schemes~\footnote{http://www.ivl.disco.unimib.it/activities/cbir-rs/}.

The rest of the paper is organized as follows: Section~\ref{background}
reviews the most relevant visual descriptors and retrieval schemes; Section~~\ref{materials} describes the data, visual descriptors,  retrieval schemes evaluated and the experimental setup; Section~\ref{results} reports and analyzes the experimental results; finally, Section~\ref{conclusions} presents our final considerations
and discusses some new directions for our future research.

\begin{figure}[tb]
  \centering
  \includegraphics[width=0.66\textwidth]{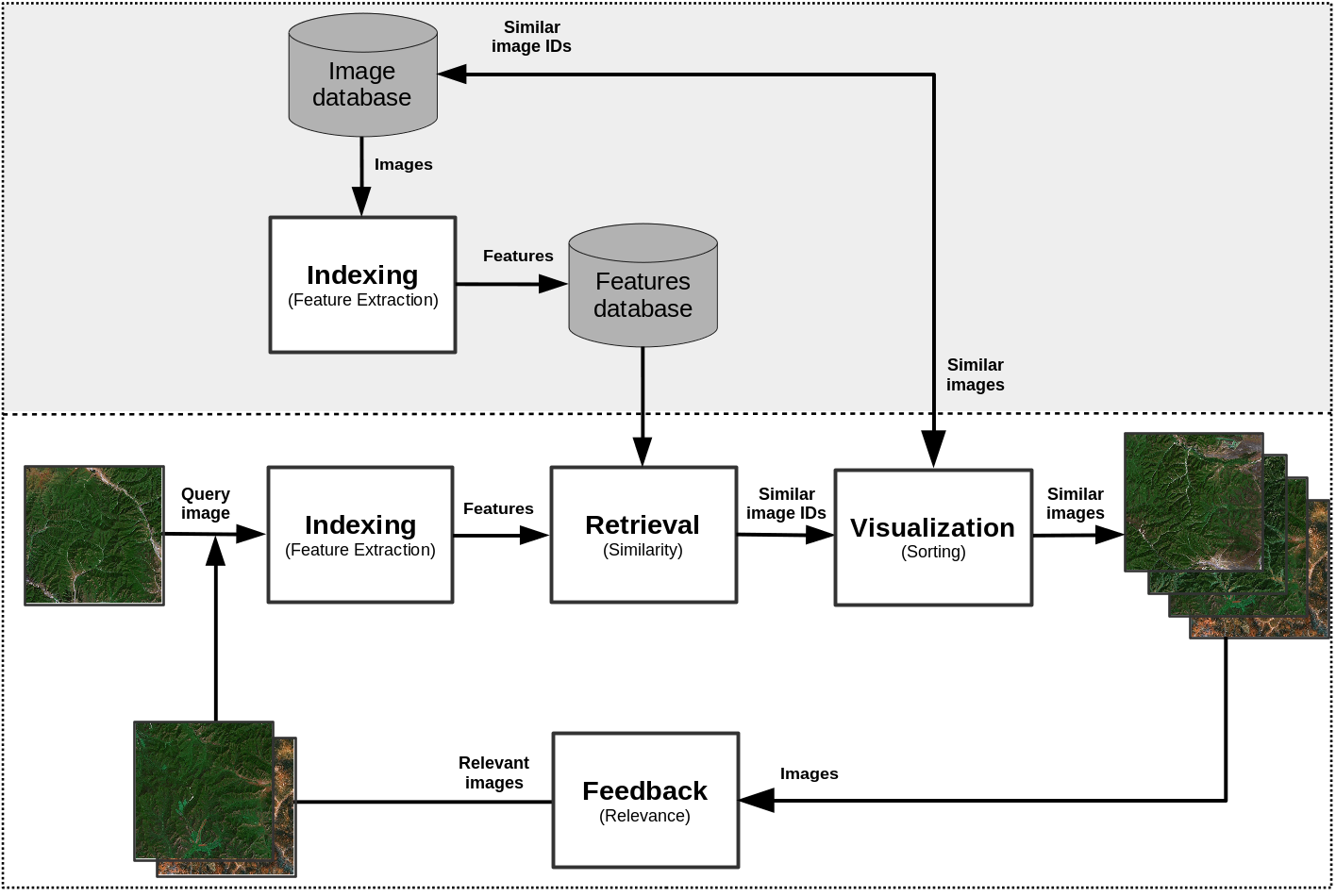} 
  \caption{Main components of a CBIR system.}
  \label{fig:CBIR}
\end{figure}

\section{Background and Related Works}\label{background}
A typical Content-Based Image Retrieval (CBIR) system is composed of four main parts~\cite{smeulders2000content,datta2008image}, see Fig.~\ref{fig:CBIR}: 
\begin{enumerate}
\item The \emph{Indexing}, also called \emph{feature extraction}, module computes the visual descriptors that characterize the image content. Given an image, these features are usually pre-computed and stored in a \emph{database of features};
\item The \emph{Retrieval} module, given a query image, finds the images in the database that are most similar by comparing the corresponding visual descriptors. 
\item The \emph{Visualization} module shows the images that are most similar to a given query image ordered by the degree of similarity.
\item The \emph{Relevance Feedback} module makes it possible to select relevant images from the subset of images returned after an initial query. This selection can be given manually by a user or automatically by the system.
\end{enumerate}

\subsection{Indexing}
\label{sec:descriptors}

A huge variety of features have been proposed in literature for describing the visual content. They are often divided into \emph{hand-crafted} features and \emph{learned} features. Hand-crafted descriptors are features extracted using a manually predefined algorithm based on the expert knowledge. 
Learned descriptors are features extracted using Convolutional Neural Networks (CNNs).

Global hand-crafted features describe an image as a whole in terms of color, texture and shape distributions~\cite{mirmehdi2009handbook}. Some notable examples of global features are color histograms~\cite{novak1992anatomy}, spatial histogram~\cite{wang2010robust}, Gabor filters~\cite{manjunath1996texture}, co-occurrence matrices~\cite{Arvis2004,haralick1979statistical}, Local Binary Patterns (LBP)~\cite{ojala2002multiresolution}, Color and Edge Directivity Descriptor (CEDD)~\cite{chatzichristofis2008cedd}, Histogram of Oriented Gradients (HOG)~\cite{junior2009trainable}, morphological operators like granulometries information~\cite{bosilj2016retrieval,aptoula2014,Kandaswamy2005}, Dual Tree Complex Wavelet Transform (DT-CWT)~\cite{Bianconi2011,Barilla2008} and GIST~\cite{oliva2001modeling}. Readers who would wish to deepen the subject can refer to the following papers~\cite{rui1999image,deselaers2008features,liu2013content,veltkamp2013state}.

Local hand-crafted descriptors like Scale Invariant Feature Transform (SIFT)~\cite{lowe2004distinctive,bianco2015local} provide a way to describe salient patches around properly chosen key points within the images. 
The dimension of the feature vector depends on the number of chosen key points in the image. A great number of key points can generate large feature vectors that can be difficult to be handled in the case of a large-scale image retrieval system. The most common approach to reduce the size of feature vectors is the \emph{Bag-of-Visual Words} (BoVW)~\cite{sivic2003video,yang2010bag}. This approach has shown excellent performance not only in image retrieval applications~\cite{deselaers2008features} but also in object recognition~\cite{grauman2010visual}, image classification~\cite{csurka2004visual} and annotation~\cite{tsai2012bag}. The idea underlying
is to quantize by clustering local descriptors into \emph{visual words}. Words are then defined as the centers of the learned clusters and are representative of several similar local regions. Given an image, for each key point the corresponding local descriptor is mapped to the most similar visual word. The final feature vector of the image is represented by the histogram of the its visual words.

CNNs are a class of learnable architectures used in many domains such as image recognition, image annotation, image retrieval etc~\cite{schmidhuber2015deep}. CNNs are usually composed of several layers of processing, each involving linear as well as non-linear operators, that are learned jointly, in an end-to-end manner, to solve a particular tasks. A typical CNN architecture for image classification consists of one or more convolutional layers followed by one or more fully connected layers. The result of the last full connected layer is the CNN output. The number of output nodes is equal to the number of image classes~\cite{krizhevsky2012imagenet}. 

A CNN that has been trained for solving a given task can be also adapted to solve a different task. In practice, very few people train an entire CNN from scratch, because it is relatively rare to have a dataset of sufficient size. Instead, it is common to take a CNN that is pre-trained on a very large dataset (e.g. ImageNet, which contains 1.2 million images with 1000 categories~\cite{deng2009imagenet}), and then use it either as an initialization or as a fixed feature extractor for the task of interest~\cite{razavian2014cnn,vedaldiCNN}. In the latter case, given an input image, the pre-trained CNN performs all the multilayered operations and the corresponding feature vector is the output of one of the fully connected layers~\cite{vedaldiCNN}. This use of CNNs have demonstrated to be very effective in many pattern recognition applications~\cite{razavian2014cnn}.

\subsection{Retrieval}

A basic retrieval scheme takes as input the visual descriptor corresponding to the query image perfomed by the user and it computes the similarity between such a descriptor and all the visual descriptors of the database of features. As a result of the search, a ranked list of images is returned to the user.  The list is ordered by a degree of similarity, that can be calculated in several ways~\cite{smeulders2000content}: Euclidean distance (that is the most used), Cosine similarity, Manhattan distance, $\chi^2$-square distance, etc.~\cite{ten2004similarity}. 

\subsection{Relevance Feedback}
In some cases visual descriptors are not able to completely represent the image semantic content. Consequently, the result of a CBIR system might be not completely satisfactory.  
One way to improve the performance is to allow the user to better specify its information need by expanding the initial query with other relevant images~\cite{rui1998relevance,hong2000incorporate,zhou2003relevance,li2013relevance}. Once the result of the initial query is available, the feedback module makes it possible to automatically or manually select a subset of relevant images. In the case of automatic relevance feedback (\emph{pseudo-relevance feedback})~\cite{baeza1999modern}, the top $n$ images retrieved are considered relevant and used to expand the query. In the case of manual relevance feedback  (\emph{explicit relevance feedback (RF)})~\cite{baeza1999modern}, it is the user that manually selects $n$ relevant of images from the results of the initial query. In both cases, the relevance feedback process can be iterated several times to better capture the information need. Given the initial query image and the set of relevant images, whatever they are selected, the feature extraction module computes the corresponding visual descriptors and the corresponding queries are performed individually. The final set of images is then obtained by combining the ranked sets of images that are retrieved. There are several alternative ways in which the relevance feedback could implemented to expand the initial query. Readers who would wish to deepen on this topic can refer the following papers~\cite{zhou2003relevance,li2013relevance,rui2001relevance}. 

The performance of the system when relevance feedback is used, strongly depends on the quality of the results achieved after the initial query. A system using effective features for indexing, returns a high number of relevant images in the first ranking positions. This  makes the pseudo-relevance feedback effective and, in the case of manual relevance feedback, it makes easier to the user selecting relevant images within the result set. 

Although there are several examples in the literature of manual RF~\cite{thomee2012interactive,ciocca1999relevance,ciocca2001quicklook}, since human labeling task is enormously boring and time consuming, these schemes are not practical and efficient in a real scenario, especially when huge archives of images are considered. Apart from the pseudo-RF, other alternatives to manual RF approach are the hybrid systems such as the systems based on supervised machine learning~\cite{Bruzzone,pedronette2015semi}. This learning method aims at finding the most informative images in the archive that, when annotated and included in the set of relevant and irrelevant images (i.e., the training set), can significantly improve the retrieval performance \cite{Bruzzone,ferecatu2007interactive}. The \emph{Active-Learning-based RF} scheme presented by Demir et al.~\cite{Bruzzone} is an example of hybrid scheme.   Given a query, the user selects a small number of relevant and not relevant images that are used as training examples to train a binary classifier based on Support Vector Machines. The system iteratively proposes images to the user that assigns the relevance feedback. At each RF iteration the classifier is re-trained using a set of images composed of the initial images and
the images from the relevance feedback provided by the user. After some RF iterations, the classifier is able to retrieve images that are similar to the query with a higher accuracy with respect to the initial query. At each RF iteration, the system suggests images to the user by following this strategy: 1) the system selects the $p$ most uncertain (i.e. ambiguous) images by taking the ones closest to the classifier hyperplanes; 2) the system selects the $h$ (with $h<p$) most diverse images from the highest density regions of the future space. 

\section{Methods and materials}\label{materials}

  Given an image database $\mathbf{D}$ composed of $M$ images, the most $k$ relevant images of $\mathbf{D}$ to a given query are the $k$ images that have the smallest distances between their feature vectors and the feature vector extracted from the query image. Let us consider $\mathbf{x}$ and  $\mathbf{y}$ as the feature vectors extracted from the query image and a generic image of $\mathbf{D}$  respectively. The distance  $d(\mathbf{x},\mathbf{y})$ between two vectors can be calculated by using several distance functions, here we considered: Euclidean, Cosine, Manhattan, and $\chi^{2}$-square. 

In this work we evaluated:
\begin{enumerate}
\item several visual descriptors as described in Sect.~\ref{descriptors};
\item different retrieval schemes as described in Sect.~\ref{schamas};
\end{enumerate}

We conducted all the experiments on two publicly available datasets described in Sec.~\ref{datars} for which the ground truth is known. 

\subsection{Visual descriptors}
\label{descriptors}

In this work we  compared visual descriptors for content-based retrieval of remote sensing images. We considered  a few representative descriptors selected from global and local hand-crafted and Convolutional Neural Networks approaches. In some cases we considered both color and gray-scale
images. The gray-scale image $L$ is defined as follows:
$L = 0.299R + 0.587G + 0.114B$. All feature vectors have been $L^2$ normalized (they have been divided by its $L^2$-norm):

\subsubsection{Global hand-crafted descriptors}
\begin{itemize}
\item 256-dimensional gray-scale histogram (Hist L)~\cite{novak1992anatomy};
\item 512-dimensional Hue and Value marginal histogram obtained from
  the HSV color representation of the image (Hist H V)~\cite{novak1992anatomy};
\item 768-dimensional RGB and \emph{rgb} marginal
  histograms (Hist RGB and Hist \emph{rgb})~\cite{Pietikainen1996};
  \item 1536-dimensional spatial RGB histogram achieved from a RGB histogram calculated in different part of the image (Spatial Hist RGB)~\cite{novak1992anatomy};
\item 5-dimensional feature vector composed of contrast, correlation,
  energy, entropy and homogeneity extracted from the
  \emph{co-occurrence matrices} of each color
  channel (Co-occ. matr.)~\cite{Arvis2004,Parkkinen1996};
  \item 144-dimensional \emph{Color and Edge Directivity Descriptor} (CEDD) features. This descriptor uses a fuzzy version of the five digital filters proposed by the MPEG-7 Edge Histogram Descriptor (EHD), forming 6 texture areas. CEDD uses 2 fuzzy systems that map the colors of the image in a 24-color custom palette;
  \item 8-dimensional \emph{Dual Tree Complex Wavelet Transform} (DT-CWT)
 features obtained considering four scales, mean and
  standard deviation, and three color  channels (DT-CWT and DT-CWT L)~\cite{Bianconi2011,Barilla2008};
  \item 512-dimensional \emph{Gist} features obtained considering eight
  orientations and four scales for each
  channel (Gist RGB)~\cite{oliva2001modeling};
\item 32-dimensional \emph{Gabor} features composed of mean and
  standard deviation of six orientations extracted at four frequencies
  for each color channel (Gabor L and Gabor RGB)~\cite{Bianconi2011,Bianconi2007};
\item 264-dimensional \emph{opponent Gabor} feature vector extracted
  as Gabor features from several inter/intra channel combinations:
  monochrome features extracted from each channel separately and
  opponent features extracted from couples of colors at different
  frequencies (Opp. Gabor RGB)~\cite{Jain1998};
\item 580-dimensional \emph{Histogram of Oriented Gradients}
  feature vector~\cite{junior2009trainable}. Nine histograms with nine
  bins are concatenated to achieve the final feature vector  (HoG);
\item 78-dimensional feature vector obtained calculating morphological
  operators (\emph{granulometries}) at four angles and for each color
  channel (Granulometry)~\cite{Kandaswamy2005};
\item 18-dimensional \emph{Local Binary Patterns} (LBP) feature
  vector for each channel. We considered LBP applied to gray images
  and to color images represented in RGB~ \cite{maenpaa2004classification}.  We selected the LBP with a circular neighbourhood of radius 2 and 16 elements, and 18 uniform and rotation invariant patterns. We set $w = 16$ and $w = 30$ for the LandUse and SceneSat datasets respectively (LBP L and LBP RGB).
\end{itemize}

\subsubsection{Local hand-crafted descriptors}

\begin{enumerate}
\item SIFT: We considered four variants of the \emph{Bag of Visual Words} (BoVW) representation of a 128-dimensional Scale Invariant Feature Transform (SIFT) calculated on the gray-scale image. For each variant, we built a codebook of \num{1024} visual words by exploiting images from external sources. 

The four variants are:
\begin{itemize}
\item SIFT: \num{1024}-dimensional BoVW of SIFT descriptors extracted from regions at given key points chosen using the SIFT detector (SIFT);
\item Dense SIFT: \num{1024}-dimensional BoVW of SIFT descriptors extracted from regions at given key points chosen from a dense grid. 
\item Dense SIFT (VLAD): \num{25600}-dimensional vector of locally aggregated descriptors
  (VLAD)~\cite{cimpoi2014describing}.
\item Dense SIFT (FV):\num{40960}-dimensional Fisher's vectors (FV) of locally
  aggregated descriptors~\cite{jegou2010aggregating}.
\end{itemize}

\item LBP: We considered the \emph{Bag of Visual Words} (BoVW) representation of Local Binary Patterns descriptor calculated on each channel of the RGB color space separately and then concatenated. LBP has been extracted from regions at given key points sampled from a dense grid every 16 pixels.  We considered the LBP with a circular neighbourhood of radius 2 and 16 elements, and 18 uniform and rotation invariant patterns~\cite{cusano2015remote-sensing}. We set $w = 16$ and $w = 30$ for the LandUse and SceneSat respectively. Also in this case the codebook was built using an external dataset (Dense LBP RGB).
\end{enumerate}
\subsubsection{CNN-based descriptors}

The CNN-based features have been obtained as the intermediate
representations of deep convolutional neural networks originally
trained for scene and object recognition. The networks are used to generate a
visual descriptor by removing the final softmax nonlinearity and the
last fully-connected layer. We selected
the most representative CNN architectures in the state of the
art~\cite{vedaldiCNN,szegedy2015going,he2016deep,arandjelovic2016netvlad} by considering a different accuracy/speed
trade-off. All the CNNs have been trained on the ILSVRC-2015 dataset~\cite{ILSVRC15}
using the same protocol as in~\cite{krizhevsky2012imagenet}. In
particular we considered \num{4096}, \num{2048}, \num{1024} and
\num{128}-dimensional feature vectors as
follows~\cite{razavian2014cnn,Marmanis}:
\begin{itemize}
\item \emph{BVLC AlexNet} (BVLC AlexNet): this is the AlexNet trained on ILSVRC
  2012~\cite{krizhevsky2012imagenet}.
\item \emph{BVLC Reference CaffeNet} (BVLC Ref): a AlexNet trained on
  ILSVRC 2012, with a minor variation~\cite{vedaldiCNN} from the version as described
  in~\cite{krizhevsky2012imagenet}.
\item \emph{Fast CNN} (Vgg F): it is similar to the one presented
  in~\cite{krizhevsky2012imagenet} with a reduced number of
  convolutional layers and the dense connectivity between
  convolutional layers. The last fully-connected layer is
  4096-dimensional~\cite{chatfield2014return}.
\item \emph{Medium CNN} (Vgg M): it is similar to the one presented
  in~\cite{zeiler2014visualizing} with a reduced number of filters in
  the convolutional layer four. The last fully-connected layer is
  4096-dimensional~\cite{chatfield2014return}.
\item \emph{Medium CNN} (Vgg M-2048-1024-128): three modifications of
  the Vgg M network, with lower dimensional last fully-connected
  layer. In particular we used a feature vector of 2048, 1024 and 128
  size~\cite{chatfield2014return}.
\item \emph{Slow CNN} (Vgg S): it is similar to the one presented
  in~\cite{sermanet2013overfeat} with a reduced number of
  convolutional layers, less filters in the layer five and the Local
  Response Normalization. The last fully-connected layer is
  4096-dimensional~\cite{chatfield2014return}.
\item \emph{Vgg Very Deep 19 and 16 layers} (Vgg VeryDeep 16 and 19):
  the configuration of these networks has been achieved by increasing
  the depth to 16 and 19 layers, that results in a substantially
  deeper network than the ones previously~\cite{simonyan2014very}.
   \item \emph{GoogleNet}~\cite{szegedy2015going} is a 22 layers deep network architecture that has been designed to improve the utilization of the computing resources inside the network.
  \item \emph{ResNet 50} is Residual Network. Residual learning framework are designed to ease the training of networks that are substantially deeper than those used previously. This network has 50 layers~\cite{he2016deep}.
  \item \emph{ResNet 101} is Residual Network made of 101 layers~\cite{he2016deep}.
  \item \emph{ResNet 152} is Residual Network made of 101 layers~\cite{he2016deep}.
\end{itemize}
Besides traditional CNN architectures, we evaluated the \emph{NetVLAD}~\cite{arandjelovic2016netvlad}. This architecture is a combination of a Vgg VeryDeep 16~\cite{simonyan2014very} and  a VLAD layer~\cite{delhumeau2013revisiting}. The network has been trained for place recognition using a subset of a large dataset of multiple panoramic images depicting the same place from different viewpoints over time from the Google Street View Time Machine~\cite{torii2013visual}. 

To further evaluate the power of CNN-based descriptors, we have fine-tuned a CNN to the remote sensing domain. We have chosen the ResNet-50 which represents a good trade-off between depth and performance. This CNN demonstrated to be very effective on the ILSVRC 2015 (ImageNet Large Scale Visual Recognition Challenge) validation set with a top 1- recognition accuracy of about 80\%~\cite{he2016deep}. 

For the fine-tuning procedure we considered a very recent RS database ~\cite{xia2017aid}, named AID, that is made of aerial image dataset collected from Google Earth imagery. This dataset is made up of the following 30 aerial scene types: airport, bare land, baseball field, beach, bridge, center, church, commercial, dense residential, desert, farmland, forest, industrial, meadow, medium residential, mountain, park, parking, playground, pond, port, railway station, resort, river, school, sparse residential, square, stadium, storage tanks and viaduct. The AID dataset has a number of 10000 images within 30 classes and about 200 to 400 samples of size 600$\times$ 600 in each class.

We did not train the ResNet-50 from the scratch on AID because the number of images for each class is not enough. We started from the pre-trained ResNet-50 on ILSVRC2015 scene image classification dataset~\cite{ILSVRC15}. From the AID dataset we have selected 20 images for each class and the rest has been using for training. During the fine-tuning stage each image has been resized to $256 \times 256$ and a random crop has been taken of $224 \times 224$ size. We augmented data with the horizontal flipping. During the test stage we considered a single central $224 \times 224$ crop from the  $256 \times 256$-resized image.

The ResNet-50 has been trained via stochastic gradient descent with a mini-batch of 16 images. We set the initial learning rate to 0.001 with learning rate update at every 2K iterations. The network has been trained within the Caffe~\cite{jia2014caffe} framework on a PC equipped with a Tesla NVIDIA K40 GPU. The classification accuracy of the resulting SatResNet-50 fine-tuned with the AID dataset is 96.34\% for the Top-1, and 99.34\% for the Top-5.

In the following experiments, the SatResNet-50 is then used as feature extractor. The activations of the neurons in the fully connected layer are used as features for the retrieval of food images. The resulting feature vectors have size 2048 components.

\subsection{Retrieval schemes}
\label{schamas}
We evaluated and compared three retrieval schemes exploiting different distance functions, namely Euclidean, Cosine, Manhattan, and $\chi^2$-square and an active-learning-based RF scheme using the histogram intersection as distance measure. In particular, we considered:

\begin{enumerate}
\item \emph{A basic IR}. This scheme takes a query as input and outputs a list of ranked similar images. 
\item \emph{Pseudo-RF}. This scheme considers the first $n$ images returned after the initial query as relevant. We considered different values of $n$ ranging between 1 and 10.
\item \emph{Manual RF}. Since the ground truth is known, we simulated the human interaction by taking the first $n$ actual relevant images from the result set obtained after the initial query. We evaluated performance at different values of $n$ ranging between 1 and 10. 
\item \emph{Active-Learning-based RF}. We considered an Active-Learning-based RF scheme as presented by Demir et al.~\cite{Bruzzone}. The RF scheme requires the interaction with the user that we simulated taking relevant and not relevant images from the ground-truth.
\end{enumerate}

\subsection{Remote Sensing Datasets}
\label{datars}

\begin{figure}[tb]
  \centering
  \setlength{\tabcolsep}{1.0pt}
  \def\arraystretch{1.0}%
  \begin{tabular}{ccccccccccc}
  \includegraphics[width=0.08\textwidth]{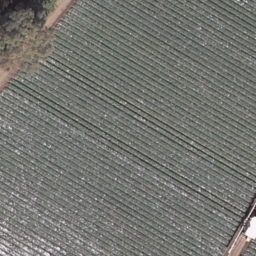} &
  \includegraphics[width=0.08\textwidth]{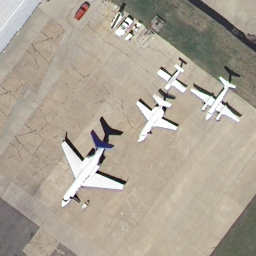} &
  \includegraphics[width=0.08\textwidth]{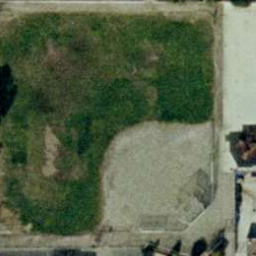} &
    \includegraphics[width=0.08\textwidth]{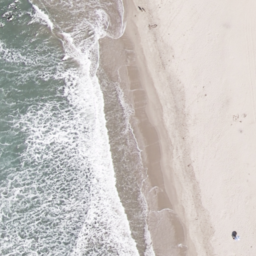} &
      \includegraphics[width=0.08\textwidth]{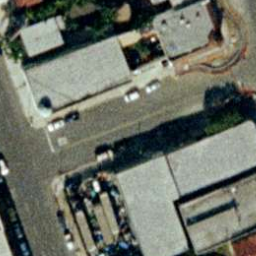} &
        \includegraphics[width=0.08\textwidth]{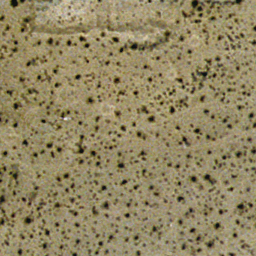} &
  \includegraphics[width=0.08\textwidth]{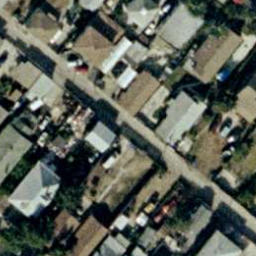}\\
   \includegraphics[width=0.08\textwidth]{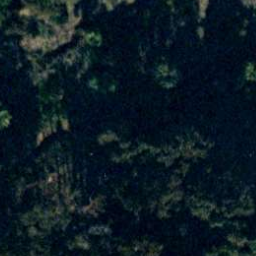} &
  \includegraphics[width=0.08\textwidth]{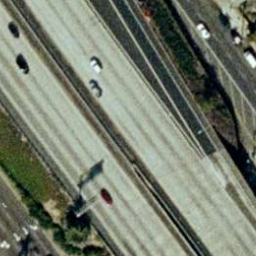} &
  \includegraphics[width=0.08\textwidth]{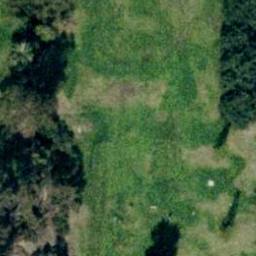} &
    \includegraphics[width=0.08\textwidth]{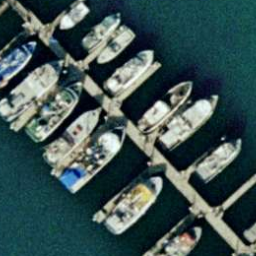}&
      \includegraphics[width=0.08\textwidth]{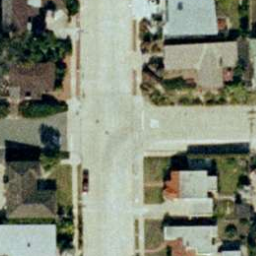} &
        \includegraphics[width=0.08\textwidth]{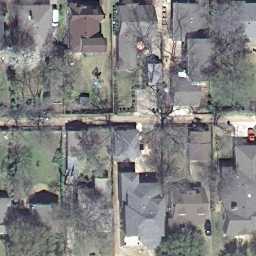} &
  \includegraphics[width=0.08\textwidth]{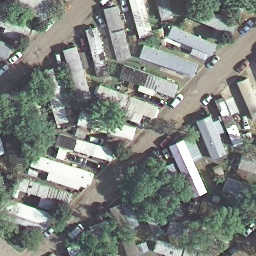} \\
     \includegraphics[width=0.08\textwidth]{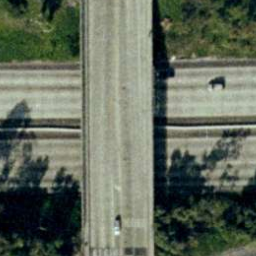} &
  \includegraphics[width=0.08\textwidth]{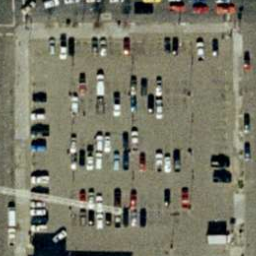} &
  \includegraphics[width=0.08\textwidth]{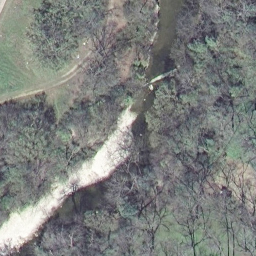} &
    \includegraphics[width=0.08\textwidth]{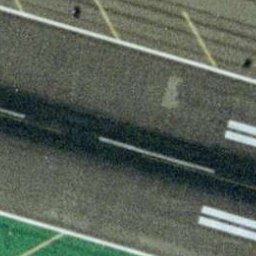} &
      \includegraphics[width=0.08\textwidth]{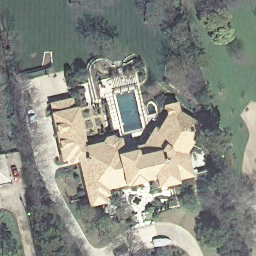} &
        \includegraphics[width=0.08\textwidth]{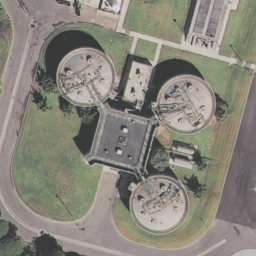} &
  \includegraphics[width=0.08\textwidth]{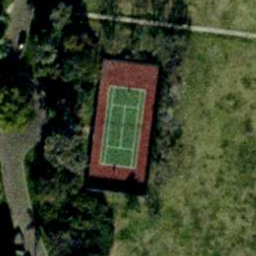} \\
   \end{tabular}
  \caption{Examples from the 21-Class Land-Use/Land-Cover dataset. From the top left to bottom right: \emph{agricultural}, \emph{airplane}, %
\emph{baseball diamond}, \emph{beach}, \emph{buildings},
\emph{chaparral}, \emph{dense residential}, \emph{forest},
\emph{freeway}, \emph{golf course}, \emph{harbor},
\emph{intersectionv}, \emph{medium density residential}, \emph{mobile
  home park}, \emph{overpass}, \emph{parking lot}, \emph{river},
\emph{runway}, \emph{sparse residential}, \emph{storage tanks}, and
\emph{tennis courts}.}
  \label{fig:landuse_dataset}
\end{figure}

\begin{figure}[tb]
  \centering
  \setlength{\tabcolsep}{1.0pt}
  \def\arraystretch{1.0}%
  \begin{tabular}{ccccccc}
  \includegraphics[width=0.08\textwidth]{airport_01} &
  \includegraphics[width=0.08\textwidth]{beach-01} &
  \includegraphics[width=0.08\textwidth]{bridge_01} &
    \includegraphics[width=0.08\textwidth]{commercial_01} &
      \includegraphics[width=0.08\textwidth]{Desert_01} &
        \includegraphics[width=0.08\textwidth]{Farmland-01} &
  \includegraphics[width=0.08\textwidth]{footballField_01} \\
   \includegraphics[width=0.08\textwidth]{forest_01}&
  \includegraphics[width=0.08\textwidth]{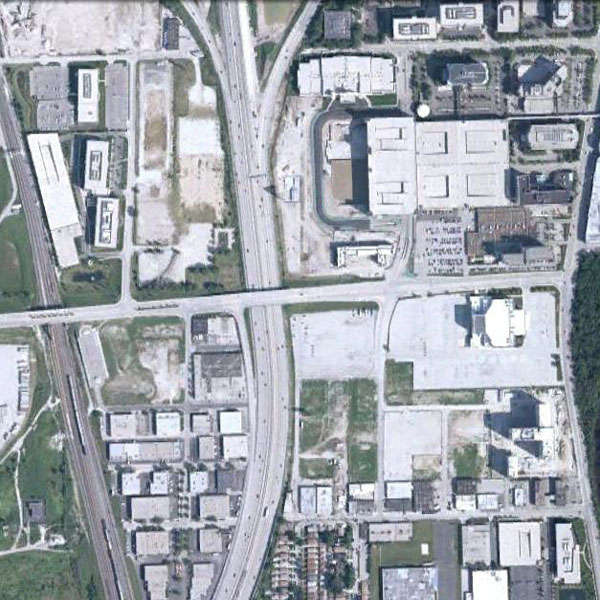} &
  \includegraphics[width=0.08\textwidth]{meadow_01} &
    \includegraphics[width=0.08\textwidth]{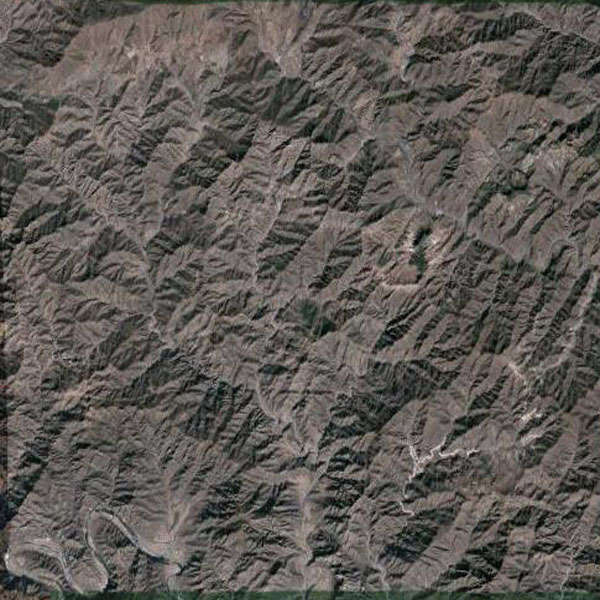} &
      \includegraphics[width=0.08\textwidth]{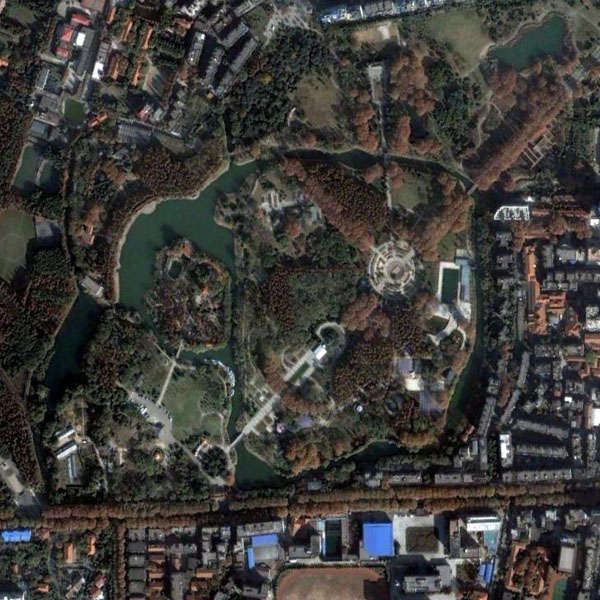} &
        \includegraphics[width=0.08\textwidth]{parking_01} &
  \includegraphics[width=0.08\textwidth]{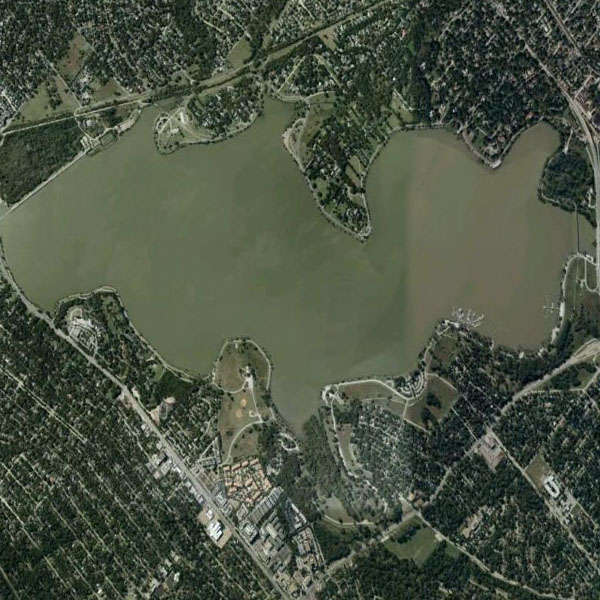} \\
     \includegraphics[width=0.08\textwidth]{port_01} &
  \includegraphics[width=0.08\textwidth]{railwayStation_01} &
  \includegraphics[width=0.08\textwidth]{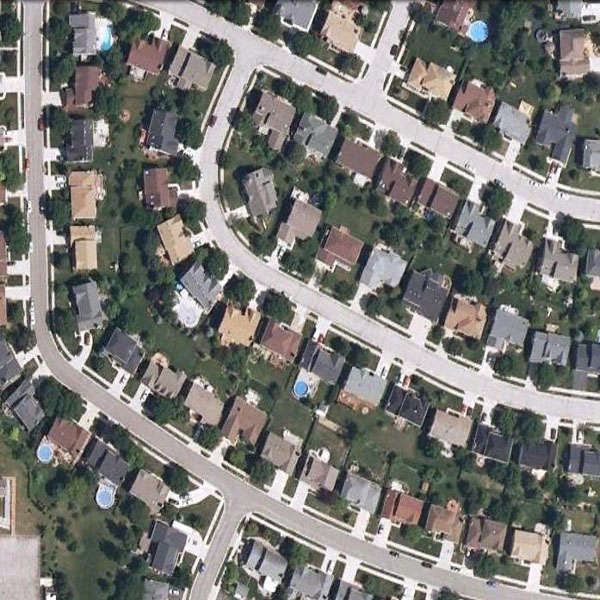} &
    \includegraphics[width=0.08\textwidth]{river_01} &
      \includegraphics[width=0.08\textwidth]{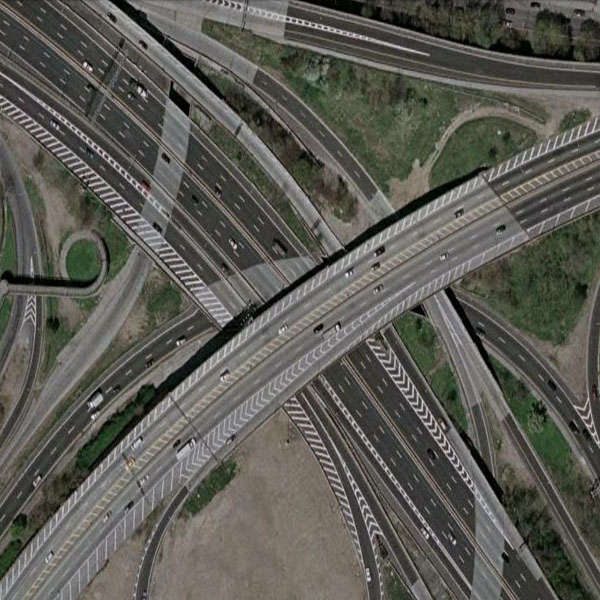} &
            \\
   \end{tabular}
  \caption{Examples from the 19-Class Satellite Scene dataset. From the top left to bottom right: \emph{airport}, \emph{beach}, \emph{bridge},
\emph{commercial area}, \emph{desert}, \emph{farmland}, \emph{football
  field}, \emph{forest}, \emph{industrial area}, \emph{meadow},
\emph{mountain}, \emph{park}, \emph{parking}, \emph{pond},
\emph{port}, \emph{railway station}, \emph{residential area},
\emph{river} and \emph{viaduct}.}
  \label{fig:rsdataset}
\end{figure}

The 21-Class Land Use/Land Cover Dataset (LandUse) is a dataset of images
of 21 land-use classes selected from aerial orthoimagery with a pixel
resolution of 30 cm~\cite{yang2010bag}. The images were downloaded
from the United States Geological Survey (USGS) National Map of some
US regions.~\footnote{\url{http://vision.ucmerced.edu/datasets}}.  
For each class, one hundred $256 \times 256$ RGB images are available for a total of 2100 images. The list of
21 classes is the following: \emph{agricultural}, \emph{airplane},
\emph{baseball diamond}, \emph{beach}, \emph{buildings},
\emph{chaparral}, \emph{dense residential}, \emph{forest},
\emph{freeway}, \emph{golf course}, \emph{harbor},
\emph{intersectionv}, \emph{medium density residential}, \emph{mobile
  home park}, \emph{overpass}, \emph{parking lot}, \emph{river},
\emph{runway}, \emph{sparse residential}, \emph{storage tanks}, and
\emph{tennis courts}. 
Some examples are shown in Fig.~\ref{fig:landuse_dataset}.

The 19-Class Satellite Scene (SceneSat) dataset consists of 19 classes of satellite scenes collected from Google Earth
(Google Inc.)~\footnote{\url{http://dsp.whu.edu.cn/cn/staff/yw/HRSscene.html}}. Each class has about fifty $600 \times 600$ RGB images for a total of 1005 images ~\cite{dai2011satellite,xia2010structural}. 
following: \emph{airport}, \emph{beach}, \emph{bridge},
\emph{commercial area}, \emph{desert}, \emph{farmland}, \emph{football
  field}, \emph{forest}, \emph{industrial area}, \emph{meadow},
\emph{mountain}, \emph{park}, \emph{parking}, \emph{pond},
\emph{port}, \emph{railway station}, \emph{residential area},
\emph{river} and \emph{viaduct}.  
An example of each class is shown in Fig.~\ref{fig:rsdataset}. 

\subsubsection{Differences between LandUse and SceneSat}
\label{sec:dbdifference}

\begin{figure*}[tb]
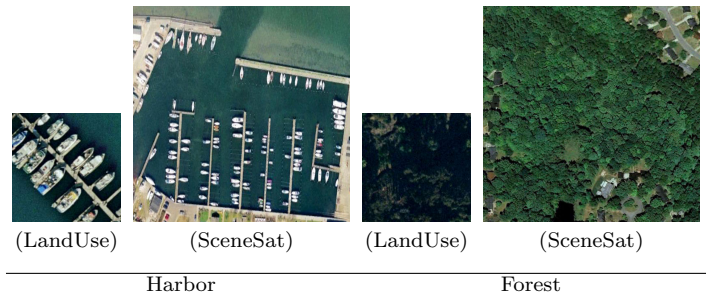

\scriptsize
  \centering
  \setlength{\tabcolsep}{2.5pt}
  \def\arraystretch{1.0}%
  \begin{tabular}{cccc}
  \includegraphics[width=0.100\textwidth]{harbor00} &
  \includegraphics[width=0.2\textwidth]{port_01} &
     \includegraphics[width=0.100\textwidth]{forest00} &
  \includegraphics[width=0.2\textwidth]{forest_01} \\
   (LandUse)& (SceneSat)&(LandUse)& (SceneSat)\\[5pt]
\bottomrule
  \multicolumn{2}{c}{Harbor}& \multicolumn{2}{c}{Forest}\\
   \end{tabular}
  \caption{Comparison between images of the same classes between LandUse and SceneSat dataset.}
  \label{fig:rsdataset_comparison}
\end{figure*}

The datasets used for the evaluation are quite different in terms of image size and resolution. LandUse images are of size $256\times256$ pixels while SceneSat images are of size $600\times600$ pixels. Fig.~\ref{fig:rsdataset_comparison} displays some images from the same category taken from the two datasets. It is quite evident that the images taken from the LandUse dataset are at a different  zoom level with respect to the images taken from the SatScene dataset. It means that objects in the LandUse dataset will be more easily recognizable than the objects contained in the SceneSat dataset, see the samples of \emph{harbour} category in Fig.~\ref{fig:rsdataset_comparison}. The SceneSat images depict a larger land area than LandUse images. It means that the SceneSat images have a more heterogeneous content than LandUse images, see some samples from \emph{harbour}, \emph{residential area} and \emph{parking} categories reported in Fig.~\ref{fig:landuse_dataset} and Fig.~\ref{fig:rsdataset}. Due to these differences between the two considered datasets, we may expect that the same visual descriptors will have different performance across datasets, see Sec.~\ref{results}.

\subsection{Retrieval measures}

Image retrieval performance has been assessed by using three state of the art measures: the \emph{Average Normalized Modified Retrieval Rank} (ANMRR), \emph{Precision} (Pr) and \emph{Recall} (Re), \emph{Mean Average Precision} (MAP)~\cite{manning2008,manjunath2001color}. We also adopted the \emph{Equivalent Query Cost} (EQC) to measure the cost of making a query independently of the computer architecture.

\subsubsection{Average Normalized Modified Retrieval Rank (ANMRR)}
The ANMRR  measure is the MPEG-7 retrieval effectiveness measure commonly accepted by the CBIR community~\cite{manjunath2001color} and largely used by recent works on content-based remote sensing image retrieval~\cite{ozkan2014,aptoula2014,yang2013}.
This metric considers the number and rank of the  relevant (ground truth) items that appear in the top images retrieved. This measure overcomes the problem related to queries with varying ground-truth set sizes. The ANMRR ranges in value between zero to one with lower values indicating better retrieval performance and is defined as follows:
$$ANMRR = \frac{1}{NQ}\sum_{q=1}^{NQ}\frac{AVR(q)-0.5[1+NG(q)]}{1.25K(q)-0.5[1+NG(q)}.$$ 

\noindent $NQ$ indicates the number of queries $q$ performed. $NG(q)$ is the size of ground-truth set for each query $q$.  $K(q)$ is a constant penalty that is assigned to items with a higher rank. $K(q)$ is commonly chosen to be $2NG(q)$. AVR is the \emph{Average Rank} for a single query $q$ and is defined as
  
  $$AVR(q) = \frac{1}{NG(q)}\sum_{k=1}^{NG(q)}Rank(k).$$  
 
\noindent where  $Rank(k)$ is the $k$th position at which a ground-truth item is retrieved. $Rank(k)$ is defined as: 
$$Rank(k) = \begin{cases}
   Rank(k), \text{if} Rank(k)\le K(q)  \\  1.25 K(q), \text{if} Rank(k)> K(q).
   \end{cases}$$

\subsubsection{Precision and Recall}
Precision is the fraction of the images retrieved that are relevant to the query

$$Pr=\frac{|\{\mbox{\# relevant images}\}\cap\{\mbox{ \# retrieved images}\}|}{|\{\mbox{\#  retrieved images}\}|}$$

It is often evaluated at a given cut-off rank, considering only the topmost $k$ results returned by the system. This measure is called \emph{precision} at $k$ or $Pr@k$.

Recall is the fraction of the images that are relevant to the query that are successfully retrieved:

$$\mbox{Re}=\frac{|\{\mbox{\# relevant images}\}\cap\{\mbox{\# retrieved images}\}|}{|\{\mbox{\# relevant images}\}|}$$

In a ranked retrieval context, precision and recall values can be plotted to give the \emph{interpolated precision-recall} curve~\cite{manning2008}. This curve is obtained by plotting the interpolated precision measured at the 11 recall levels of 0.0, 0.1, 0.2, ..., 1.0. The interpolated precision  $P_{interp}$ at a certain recall level $k$ is defined as the highest precision found for any recall level $k' \ge k$:

$$P_{interp}(k) = \max_{k' \ge k} P(k')$$

\subsubsection{Mean Average Precision (MAP)}
Given a set of queries, \emph{Mean Average Precision} is defined as,
$$MAP = \frac{\sum_{q=1}^{Q} AvePr(q)}{Q}.$$
\noindent where the average precision $AvePr$ for each query $q$ is defined as,
 
$$AvePr = \frac{\sum_{k=1}^n (Pr(k) \times rel(k))}{\mbox{\# of relevant images}} $$
where $k$ is the rank in the sequence of retrieved images, $n$ is the number of retrieved images, $Pr(k)$ is the precision at cut-off $k$  in the list ($Pr@k$), and $rel(k)$ is an indicator function equalling 1 if the item at rank $k$ is a relevant image, zero otherwise.

\subsubsection{Equivalent Query Cost}

Several previous works, such as \cite{aptoula2014}, report a table that compares the computational time needed to execute a query when different indexing techniques have been used. This comparison can not be replicated because the computational time strictly depends on the computer architecture.
To overcome this problem, we defined the \emph{Equivalent Query Cost} ($EQC$) that measures the computational cost needed to execute a given query independently of the computer architecture.  
This measure is based on the fact that the calculation of the distance between two visual descriptors is linear in number of components and on the definition of \emph{basic cost} $C$. The \emph{basic cost} is defined as the amount of computational effort that is needed to execute a single query over the entire database $\mathbf{D}$ when a visual descriptor of length $B$ is used as indexing technique. The $EQC$ of a generic visual descriptor of length $L$ can be obtained as follows:
$$EQC = C  \floor*{\frac{L}{B}},$$
where the symbol $\floor*{x}$ stands for the integer part of the number $x$, while $B$ is set to 5, that corresponds to the length of the co-occurrence matrices, that is the shortest descriptor evaluated in the experiments presented in this work.

\section{Results}\label{results}

\begin{table}[tb]
  \centering
     \resizebox{0.7\textwidth}{!}{

  \begin{tabular}{lcccccccc}
    \toprule
 Features 	&	 ANMRR	&	 MAP	&	 P@5		& P@10	&	 P@50	&	 P@100	&	 P@1000 & EQC	\\
 \bottomrule
  \emph{Global}&&&&&&&&  \\
                         Hist. L & 	0.816 & 	12.46 & 	36.65 & 	30.17 & 	18.07 & 	13.74 & 	5.96 & 	51 \\
                              Hist. H V  & 	0.781 & 	15.98 & 	54.22 & 	43.49 & 	23.41 & 	16.84 & 	6.27 & 	102 \\
                               Hist. RGB & 	0.786 & 	15.39 & 	51.82 & 	41.83 & 	22.29 & 	16.35 & 	6.14 & 	153 \\
                        Hist. \emph{rgb} & 	0.800 & 	14.34 & 	49.46 & 	38.97 & 	20.88 & 	15.31 & 	6.00 & 	153 \\
                       Spatial Hist. RGB & 	0.808 & 	14.36 & 	37.70 & 	31.13 & 	19.09 & 	14.62 & 	5.95 & 	307 \\
                            Co-occ. matr. & 	0.861 & 	8.69 & 	19.36 & 	17.20 & 	12.14 & 	10.06 & 	5.74 & 	1 \\
                                    CEDD & 	0.736 & 	19.89 & 	62.45 & 	52.39 & 	29.54 & 	20.86 & 	6.49 & 	28 \\
                                DT-CWT L & 	0.707 & 	21.04 & 	39.64 & 	36.36 & 	26.81 & 	22.11 & 	7.90 & 	1 \\
                                  DT-CWT & 	0.676 & 	24.53 & 	55.63 & 	48.92 & 	32.52 & 	25.09 & 	7.95 & 	4 \\
                                Gist RGB & 	0.781 & 	17.65 & 	45.94 & 	38.97 & 	23.10 & 	17.01 & 	6.09 & 	102 \\
                                 Gabor L & 	0.766 & 	16.08 & 	44.60 & 	37.28 & 	22.65 & 	17.63 & 	7.11 & 	6 \\	
                               Gabor RGB & 	0.749 & 	18.06 & 	52.72 & 	44.48 & 	25.71 & 	19.13 & 	7.04 & 	19 \\
                          Opp. Gabor RGB & 	0.744 & 	18.76 & 	53.81 & 	44.89 & 	26.18 & 	19.69 & 	6.99 & 	52 \\
                                     HoG & 	0.751 & 	17.85 & 	48.67 & 	41.88 & 	25.37 & 	19.12 & 	6.18 & 	116 \\
                            Granulometry & 	0.779 & 	15.45 & 	39.36 & 	33.31 & 	20.76 & 	16.30 & 	7.15 & 	15 \\
                                   LBP L & 	0.760 & 	16.82 & 	52.77 & 	45.16 & 	26.34 & 	18.84 & 	6.01 & 	3 \\
                                 LBP RGB & 	0.751 & 	17.96 & 	58.73 & 	49.83 & 	28.12 & 	19.62 & 	6.07 & 	10 \\
[4pt]
                                   \emph{Local hand-crafted}&&&&&&&&  \\
                          Dense LBP RGB & 	0.744 & 	19.01 & 	60.10 & 	51.89 & 	29.12 & 	20.30 & 	6.33 & 	204  \\
                        SIFT  & 	0.635 & 	28.49 & 	53.56 & 	49.40 & 	35.98 & 	28.42 & 	8.26 & 	204  \\
                       Dense SIFT & 	0.672 & 	25.44 & 	72.30 & 	62.61 & 	35.51 & 	25.96 & 	7.12 & 	204  \\
                Dense SIFT (VLAD) & 	0.649 & 	28.01 & 	74.93 & 	65.25 & 	38.20 & 	28.10 & 	7.18 & 	5120  \\
                  Dense SIFT (FV) & 	0.639 & 	29.18 & 	75.34 & 	66.28 & 	39.09 & 	28.54 & 	7.88 & 	8192  \\[4pt]
                                   \emph{CNN-based}&&&&&&&&  \\
                                   Vgg F & 	0.386 & 	53.55 & 	85.00 & 	79.73 & 	62.29 & 	50.24 & 	9.57 & 	819  \\
                                   Vgg M & 	0.378 & 54.44 & 86.16 & 	81.03 & 	63.42 & 50.96 & 	9.59 & 	819  \\
                                   Vgg S & 	0.381 & 	54.18 & 	86.10 & 	81.18 & 	63.46 & 	50.50 & 	9.60 & 	819 \\
                              Vgg M 2048 & 	0.388 & 	53.16 & 	85.04 & 	80.26 & 	62.77 & 	50.14 & 	9.52 & 	409  \\
                              Vgg M 1024 & 	0.400 & 	51.66 & 	84.43 & 	79.41 & 	61.40 & 	48.88 & 	9.50 & 	204  \\
                               Vgg M 128 & 	0.498 & 	40.94 & 	73.82 & 	68.30 & 	50.67 & 	39.92 & 	9.18 & 	25  \\
                                BVLC Ref & 	0.402 & 	52.00 & 	84.73 & 	79.37 & 	61.10 & 	48.96 & 	9.49 & 	819  \\
                            BVLC AlexNet & 	0.410 & 	51.13 & 	84.06 & 	78.68 & 	59.99 & 	48.01 & 	9.51 & 	819  \\
                         Vgg VeryDeep 16 & 	0.394 & 	52.46 & 	83.91 & 	78.34 & 	61.38 & 	49.78 & 	9.60 & 	819  \\
                         Vgg VeryDeep 19 & 	0.398 & 	51.95 & 	82.84 & 	77.60 & 	60.69 & 	49.16 & 	9.63 & 	819  \\
                                         GoogleNet & 	0.360 & 	55.86 & 	85.36 & 	80.96 & 	64.71 & 	52.36 & 	9.68 & 	204 \\
                               ResNet-50 & 	0.358 & 	56.57 & 	88.26 & 	84.00 & 	65.92 & 	52.69 & 	9.73 & 	409 \\
                              ResNet-101 & 	0.356 & 	56.63 & 	88.49 & 	83.53 & 	65.69 & 	52.83 & 	9.75 & 	409 \\
                              ResNet-152 & 	0.362 & 	56.03 & 	88.42 & 	83.08 & 	64.65 & 	52.50 & 	9.72 & 	409 \\
                                 NetVLAD & 	0.406 & 	51.44 & 	83.00 & 	78.59 & 	61.63 & 	49.04 & 	9.29 & 	819 \\
                            SatResNet-50 & 	\bf{0.239} & \bf{	69.94} & 	\bf{92.06} & 	\bf{89.20} & 	\bf{77.23} & \bf{64.42} & \bf{9.86} & 	409 \\
                          
    \bottomrule\\
  \end{tabular}
  }
  \caption{LandUse Dataset results obtained with a basic retrieval system with the Euclidean distance. The lower is the value of $ANMRR$ and $EQC$ the better is the performance. For the other metrics is the opposite. The best result is reported in bold.}
  \label{tab:land_basic}
\end{table}

\begin{table}[tb]
  \centering
  \resizebox{0.7\textwidth}{!}{%
  \begin{tabular}{lccccccccc}
    \toprule
 Features 	&	 ANMRR	&	 MAP	&	 P@5		& P@10	&	 P@50	&	 P@100&P@1000& EQC	\\
 \bottomrule
                                   \emph{Global hand-crafted}&&&&&&&&  \\
           Hist. L & 	0.728 & 	19.86 & 	37.69 & 	32.61 & 	21.05 & 	15.98 & 	5.21 & 	51  \\
                              Hist. H V  & 	0.704 & 	23.23 & 	43.98 & 	37.29 & 	23.10 & 	17.05 & 	5.21 & 	102  \\
                               Hist. RGB & 	0.722 & 	21.24 & 	40.96 & 	34.71 & 	21.17 & 	16.30 & 	5.20 & 	153  \\
                        Hist. \emph{rgb} & 	0.702 & 	23.03 & 	43.76 & 	37.87 & 	23.31 & 	17.11 & 	5.21 & 	153  \\
                       Spatial Hist. RGB & 	0.720 & 	22.21 & 	38.85 & 	33.36 & 	21.81 & 	16.30 & 	5.21 & 	307  \\
                                                 Co-occ. matr. & 	0.822 & 	11.73 & 	21.25 & 	18.16 & 	12.94 & 	11.00 & 	5.19 & 	1  \\
                                    CEDD & 	0.684 & 	24.15 & 	38.13 & 	34.77 & 	24.65 & 	18.52 & 	5.20 & 	28  \\
                                DT-CWT L & 	0.672 & 	23.48 & 	35.90 & 	32.43 & 	24.32 & 	20.20 & 	5.21 & 	1  \\
                                  DT-CWT & 	0.581 & 	33.16 & 	51.00 & 	45.98 & 	32.99 & 	24.52 & 	5.21 & 	4  \\
                                Gist RGB & 	0.706 & 	22.98 & 	41.73 & 	37.31 & 	22.98 & 	16.81 & 	5.19 & 	102  \\
                                 Gabor L & 	0.685 & 	22.84 & 	40.82 & 	35.63 & 	23.45 & 	19.07 & 	5.21 & 	6  \\
                               Gabor RGB & 	0.649 & 	27.00 & 	49.19 & 	43.42 & 	26.92 & 	20.70 & 	5.20 & 	19  \\
                          Opp. Gabor RGB & 	0.638 & 	28.08 & 	48.14 & 	42.48 & 	28.61 & 	21.01 & 	5.20 & 	52  \\
                                     HoG & 	0.724 & 	19.97 & 	40.24 & 	35.31 & 	21.73 & 	15.82 & 	5.20 & 	16  \\
                            Granulometry & 	0.717 & 	21.41 & 	39.20 & 	33.60 & 	20.78 & 	17.22 & 	5.21 & 	15  \\
                                   LBP L & 	0.690 & 	22.61 & 	47.24 & 	40.55 & 	24.06 & 	18.16 & 	5.20 & 	48  \\
                                 LBP RGB & 	0.664 & 	24.95 & 	50.33 & 	43.98 & 	26.33 & 	19.40 & 	5.20 & 	10  \\[4pt]
                                   \emph{Local hand-crafted}&&&&&&&&  \\
                          Dense LBP RGB & 	0.660 & 	24.81 & 	51.12 & 	44.29 & 	26.55 & 	19.67 & 	5.21 & 	204  \\
                        SIFT & 	0.559 & 	35.47 & 	59.40 & 	53.22 & 	35.04 & 	25.49 & 	5.20 & 	204  \\
                       Dense SIFT & 	0.603 & 	31.29 & 	64.06 & 	55.80 & 	31.70 & 	22.24 & 	5.20 & 	204  \\
                Dense SIFT (VLAD) & 	0.552 & 	35.89 & 	71.30 & 	62.78 & 	36.19 & 	25.03 & 	5.20 & 	5120  \\
              Dense SIFT (FV) & 	0.518 & 	39.44 & 	72.34 & 	64.69 & 	38.84 & 	27.23 & 	5.20 & 	8192  \\[4pt]
                                   \emph{CNN-based}&&&&&&&&  \\
                                   Vgg F & 	0.408 & 	49.91 & 71.52 & 	68.98 & 	49.62 & 	33.07 & 	5.21 & 	819  \\
                                   Vgg M & 	0.419 & 	48.59 & 	71.50 & 	68.27 & 	48.62 & 	32.45 & 	5.21 & 	819  \\
                                   Vgg S & 	0.416 & 	48.89 & 	71.46 & 	68.62 & 	48.79 & 	32.58 & 	5.20 & 	819 \\
                              Vgg M 2048 & 	0.431 & 	47.14 & 	71.08 & 	67.52 & 	47.33 & 	31.83 & 	5.21 & 	409  \\
                              Vgg M 1024 & 	0.443 & 	45.86 & 	70.51 & 	66.61 & 	46.05 & 	31.23 & 	5.21 & 	204  \\
                               Vgg M 128 & 	0.551 & 	34.54 & 	59.30 & 	54.08 & 	36.05 & 	25.65 & 	5.20 & 	25  \\
                                BVLC Ref & 	0.407 & 	50.04 & 	71.22 & 	68.65 & 49.75 & 	33.15 & 	5.21 & 	819  \\
                            BVLC AlexNet & 	0.421 & 	48.52 & 	70.45 & 	66.91 & 	48.22 & 	32.51 & 	5.20 & 	819  \\
                         Vgg VeryDeep 16 & 	0.440 & 	46.18 & 	70.67 & 	66.71 & 	46.22 & 	31.46 & 	5.20 & 	819  \\
                         Vgg VeryDeep 19 & 	0.455 & 	44.34 & 	69.17 & 	64.65 & 	44.84 & 	30.66 & 	5.20 & 	819  \\
                                        GoogleNet & 	0.324 & 	60.36 & 	85.73 & 	82.28 & 	68.32 & 	55.75 & 	9.75 & 	204 \\
                               ResNet-50 & 	0.329 & 	60.32 & 	88.67 & 	85.51 & 	69.44 & 	55.43 & 	9.79 & 	409 \\
                              ResNet-101 & 	0.327 & 	60.37 & 	88.81 & 	85.10 & 	68.81 & 	55.63 & 	9.79 & 	409 \\
                              ResNet-152 & 	0.332 & 	59.80 & 	88.55 & 	84.93 & 	67.94 & 	55.39 & 	9.78 & 	409 \\
                                 NetVLAD & 	0.371 & 	56.37 & 	82.54 & 	78.41 & 	64.40 & 	52.19 & 	9.48 & 	819 \\
                            SatResNet-50 & 	\bf{0.207} & 	\bf{74.19} & 	\bf{92.11} & \bf{90.55} & 	\bf{80.91} & 	\bf{68.02} & 	\bf{9.87} & 	409 \\
                     
    \bottomrule\\
  \end{tabular}
  }
  \caption{SatScene dataset results obtained with a basic retrieval system with the Euclidean distance. The lower is the value of $ANMRR$ and $EQC$ the better is the performance. For the other metrics is the opposite. The best result is reported in bold.}
    \label{tab:rs_basic}
\end{table}

\begin{figure*}[tb]
  \centering
    \begin{tabular}{cc}
    \includegraphics[width=0.7\textwidth]{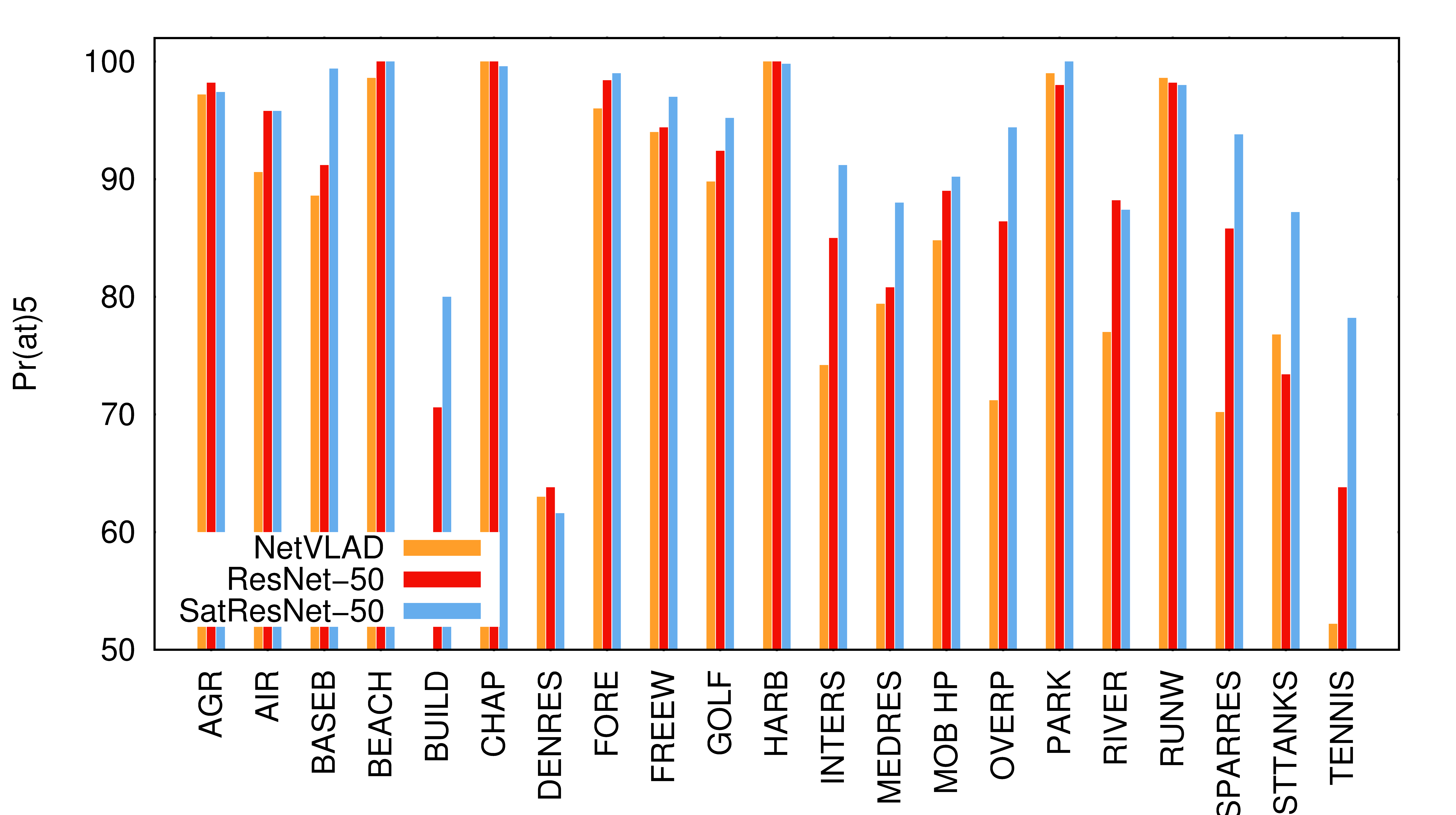}\\
    (a)\\
    \includegraphics[width=0.7\textwidth]{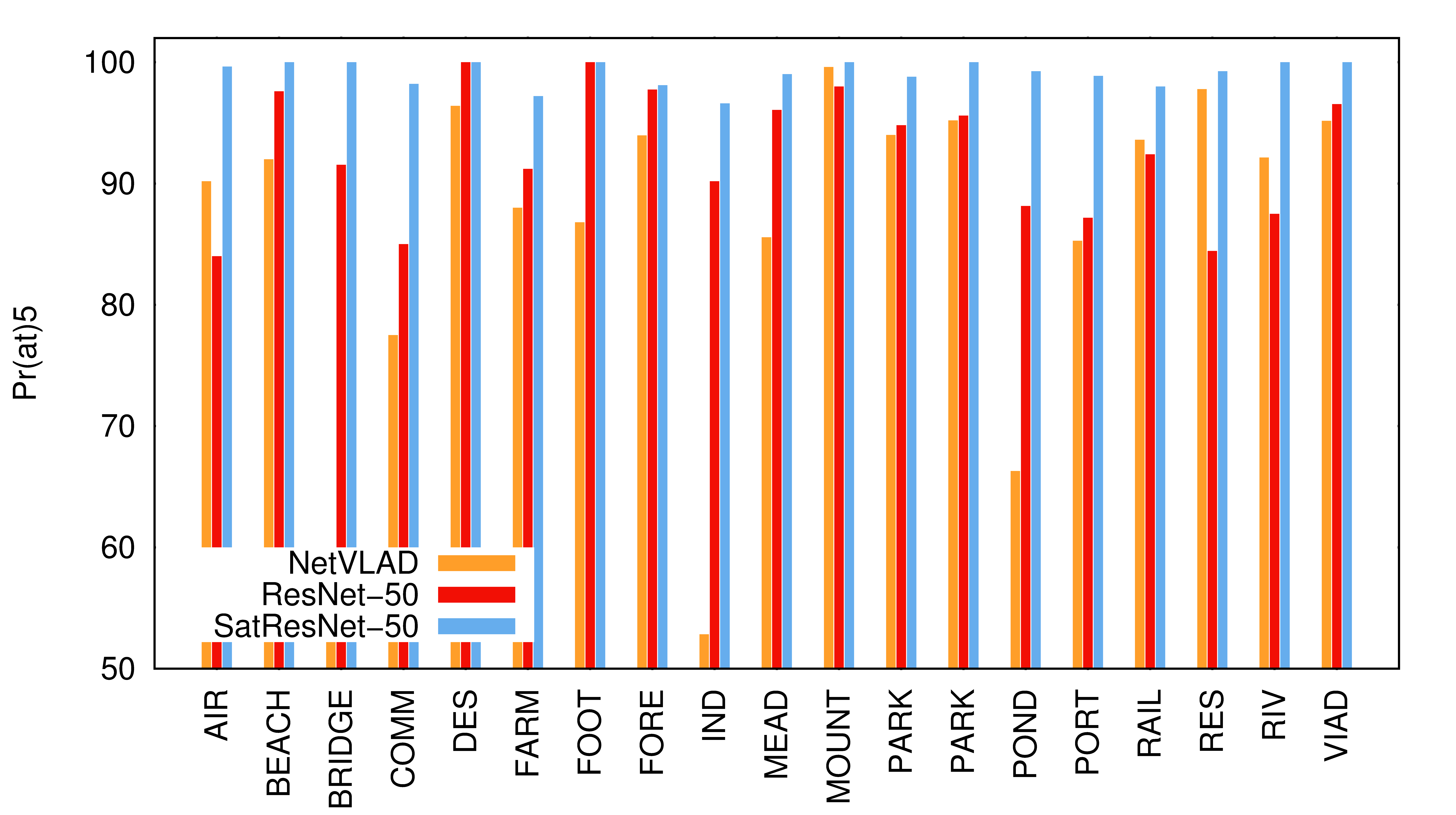}\\
        (b)\\
   \end{tabular}
  \caption{Per class precision at 5 of a selection of visual descriptors for each dataset. (a) LandUse. (b) RS.}
  \label{fig:bars}
\end{figure*}

\begin{figure*}[tb]
  \centering
    \begin{tabular}{cc}
    \includegraphics[width=0.41\textwidth]{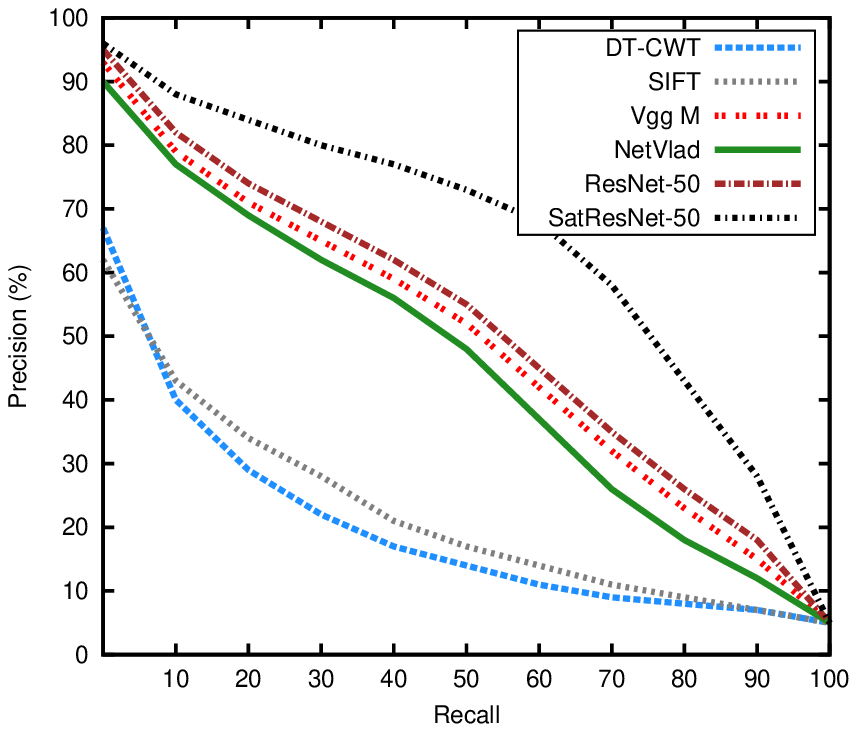}&
    \includegraphics[width=0.41\textwidth]{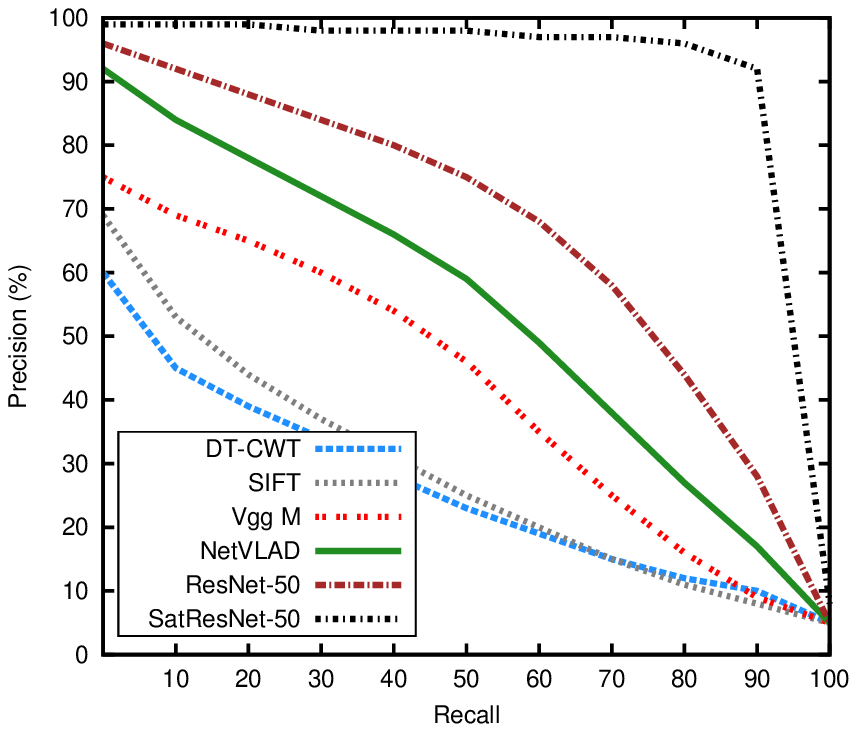}\\
    (a)&(b)\
   \end{tabular}
  \caption{Interpolated 11-points \emph{precision-recall} curves of a selection of visual descriptors for each dataset. (a) LandUse. (b) SatScene.}
  \label{fig:interpolated_pr}
\end{figure*}

\subsection{Feature evaluation using the basic retrieval scheme}

In this section we compare visual descriptors listed in Sec.~\ref{descriptors} by using the basic retrieval scheme. In order to make the results more concise, in this section we show only the experiments performed employing the Euclidean distance. Given an image dataset, in turn, we used each image as query image and evaluated the results according to the metrics discussed above, i.e.  $ANMRR$, $MAP$, $P@5$, $P@10$, $P@50$, $P@100$, $P@1000$ and $EQC$. In the case of the LandUse dataset we performed 2100 queries while in the case of SatScene dataset we evaluated 1005 queries in total.

The results obtained on the LandUse dataset are showed in Table~\ref{tab:land_basic}, while those obtained on the SatScene dataset are showed in Table~\ref{tab:rs_basic}. Regarding the LandUse dataset, the best results are obtained by using the CNN-based descriptors and in particular the ResNet CNN architectures and the SatResNet-50 that is the fine-tuned ResNet-50. The global hand-crafted descriptors have the lowest performance, with the co-occurrence matrices being the worst one. The local hand-crafted descriptors achieve better results than global hand-crafted descriptors but worse than CNN-based descriptors. In particular, the SatResNet-50, compared with Bag of Dense SIFT and DT-CWT, achieves an $ANMRR$ value that is lower of about 50\%, a $MAP$ value that is higher of about 50\% , a $P@5$ that is higher of about 50\%, a $P@10$ value that is higher of about 50\%. The same behavior can be observed for the remaining  precision levels. In particular, looking at $P@100$ we can notice that only the SatResNet-50 descriptor is capable of retrieving about 65\% of the existing images for each class ($P@100=64.42$). Regarding the SatScene dataset, the best results are obtained by the CNN-based descriptors and in particular the ResNet CNN architectures and the SatResNet-50. The global hand-crafted descriptors have the lowest performance, with the co-occurrence matrices being the worst one. The local hand-crafted descriptors achieve better results than global hand-crafted descriptors but worse than CNN-based descriptors. In particular the SatResNet-50, compared with Bag of Dense SIFT (FV), achieves an $ANMRR$ value that is lower of about 60\%, a $MAP$ value that is higher of about 50\% , a $P@5$ that is lower of about 20\%, a $P@10$ value that is higher of about 30\%. Similar behavior can be observed for the remaining  precision levels. In particular, looking at $P@50$ we can notice that only SatResNet-50 is capable of retrieving about 70\% of the existing images for each class ($P@50=68.02$).

The first columns of tables~\ref{tab:class_land_alrf} and~\ref{tab:class_rs_alrf} show the best performing visual descriptor for each remote sensing image class. For both LandUse and SceneSat datasets, the CNN-based descriptors are the best in the retrieval of all classes. SatResNet-50 performs better than other CNN architectures on most classes apart some classes containing objects rotated and translated on the image plane. In this case, NetVLAD demonstrated to perform better. Looking at Fig.~\ref{fig:bars} it is interesting to note that NetVLAD, which considers CNN features combined with local features, works better on object-based classes and more important that the SatResNet-50 network clearly outperforms the ResNet-50 thus demonstrating that the domain adaptation of the network to the remote sensing domain helped to handle with the heterogeneous content of remote sensing images.

In Fig.~\ref{fig:interpolated_pr} the \emph{interpolated 11-points precision-recall} curves achieved by a selection of  visual descriptors are plotted. It is clear that, in this experiments CNN-based descriptors outperform again other descriptors. It is interesting to note that the  SatResNet-50 network clearly outperforms the ResNet-50 thus confirming that the domain adaption has been very effective especially in the case of the SceneSat dataset. This is mostly due to the fact that both the AID and SceneSat datasets are made of pictures taken from Google Earth and then the image content is more similar. In contrast the LandUse dataset is made of picture taken from an aerial device and then the content is quite different in terms of resolution as already discussed in Section ~\ref{sec:dbdifference}.


Concerning the computational cost, the Bag Dense SIFT (FV) is the most costly solution with the worst cost-benefit trade-off. Early after the Bag Dense SIFT (FV), the Vgg M is the other most costly descriptor that is about 200 more costly than the DT-CWT, that is among the global hand-crafted descriptors the best performing one.

One may prefer a less costly retrieval strategy that is less precise and then choose for the DT-CWT. Among the CNN-based descriptors, the Vgg M 128 has better $ANMRR$ values than the DT-CWT for both datasets. The  Vgg M 128 is six times more costly than DT-CWT. Concluding, the Vgg M 128 descriptor has the best cost-benefit trade-off.

\begin{figure*}[tb]
  \centering
    \begin{tabular}{cc}
    \includegraphics[width=0.42\textwidth]{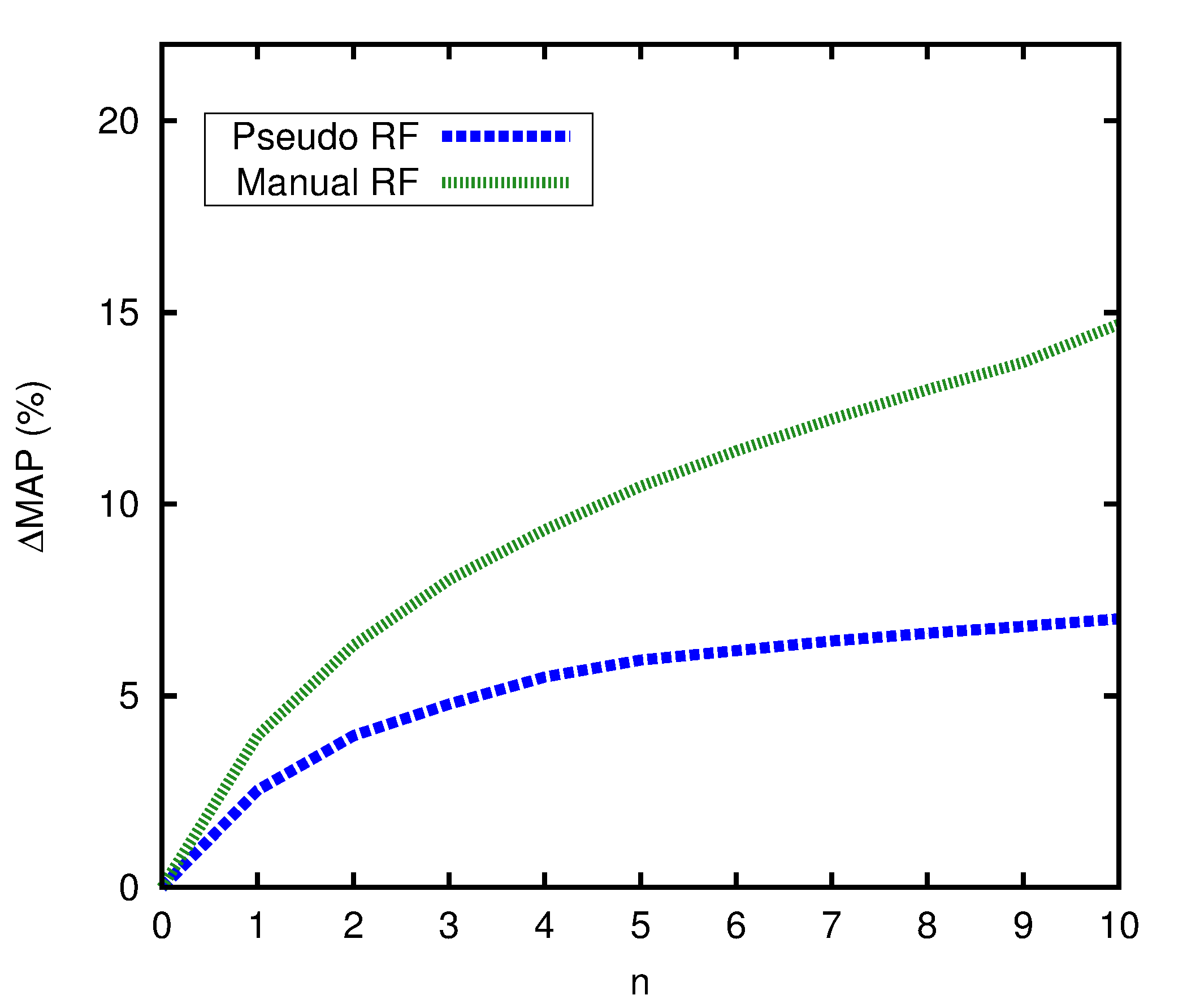}&
    \includegraphics[width=0.42\textwidth]{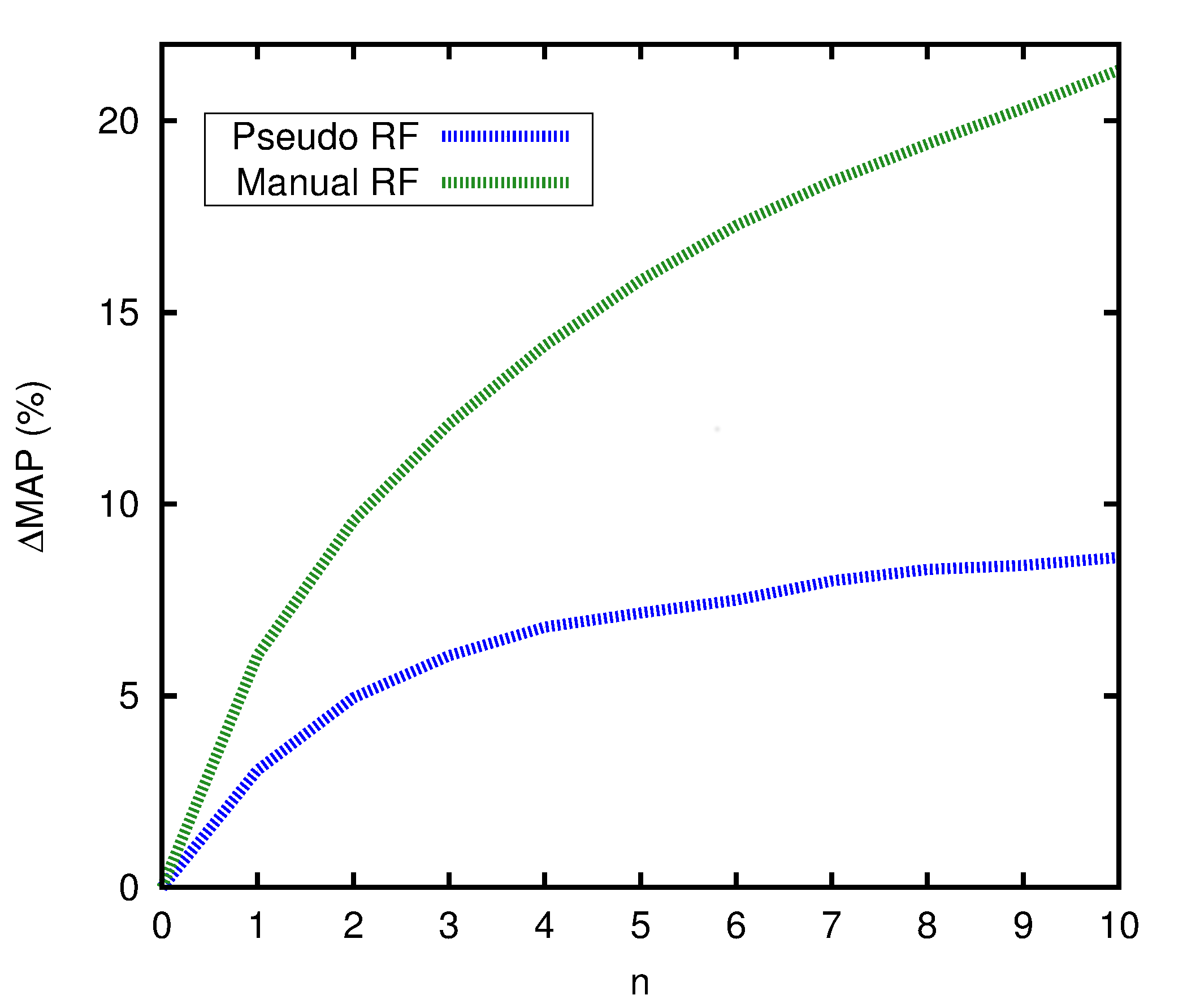}\\
    (a)&(b)\\
   \end{tabular}
  \caption{Difference of performance  ($\Delta MAP$), when the Vgg M is employed, between the pseudo RF and the basic retrieval system, and between manual RF and the basic retrieval system. (a) LandUse. (b) SceneSat.}
  \label{fig:increment}
\end{figure*}

\begin{table*}[h!]
  \setlength{\tabcolsep}{2.5pt}
  \def\arraystretch{1.0}%
 \scriptsize
      \resizebox{0.9\textwidth}{!}{
\begin{subtable}{0.5\linewidth}
\centering
\begin{tabular}{lcccccc}
    \toprule
 features 	&	 ANMRR	&	 MAP	&	 P@5		& P@10	&	 P@50	&	 P@100 	\\
 \bottomrule
                        \emph{Global hand-crafted}&&&&&&  \\
 						Hist. L & 	0.820 & 	12.31 & 	35.16 & 	28.78 & 	17.42 & 	13.43  \\ 	
                              Hist. H V  & 	0.783 & 	15.94 & 	53.16 & 	42.82 & 	23.03 & 	16.62 \\ 	
                               Hist. RGB & 	0.789 & 	15.37 & 	51.41 & 	41.22 & 	21.89 & 	16.10   \\ 	
                        Hist. \emph{rgb} & 	0.804 & 	14.16 & 	48.09 & 	37.01 & 	20.19 & 	14.93  \\ 	
                       Spatial Hist. RGB & 	0.824 & 	13.87 & 	35.15 & 	27.60 & 	16.96 & 	13.27   \\ 	
                                                 Co-occ. matr. & 	0.863 & 	8.56 & 	19.02 & 	16.68 & 	11.88 & 	9.87   \\ 	
                                    CEDD & 	0.740 & 	19.88 & 	62.27 & 	52.58 & 	29.25 & 	20.55 \\ 	
                                DT-CWT L & 	0.708 & 	21.03 & 	39.50 & 	35.67 & 	26.67 & 	22.04   \\ 	
                                  DT-CWT & 	0.677 & 	24.59 & 	55.13 & 	48.30 & 	32.24 & 	25.00   \\ 	
                                Gist RGB & 	0.804 & 	17.14 & 	44.09 & 	35.79 & 	20.46 & 	15.12 \\ 	
                                 Gabor L & 	0.769 & 	15.92 & 	44.14 & 	36.39 & 	22.16 & 	17.38 \\ 	
                               Gabor RGB & 	0.750 & 	18.04 & 	52.56 & 	44.00 & 	25.53 & 	18.97  \\ 	
                          Opp. Gabor RGB & 	0.748 & 	18.62 & 	53.73 & 	44.48 & 	25.78 & 	19.32   \\ 	
                                     HoG & 	0.757 & 	17.79 & 	47.73 & 	40.51 & 	24.69 & 	18.61\\ 	
                            Granulometry & 	0.783 & 	15.26 & 	38.54 & 	31.92 & 	20.29 & 	16.00  \\ 	
                                   LBP L & 	0.762 & 	16.82 & 	52.23 & 	44.32 & 	26.25 & 	18.61 \\ 	
                                 LBP RGB & 	0.752 & 	18.06 & 	58.35 & 	49.75 & 	28.16 & 	19.53 \\ [4pt]
                                   \emph{Local hand-crafted}&&&&&&  \\
                          Dense LBP RGB & 	0.747 & 	19.09 & 	59.49 & 	50.99 & 	28.76 & 	20.17  \\ 	
                        SIFT & 	0.648 & 	28.56 & 	52.97 & 	47.53 & 	34.65 & 	27.39   \\ 	
                       Dense SIFT & 	0.675 & 	25.67 & 	72.20 & 	62.46 & 	35.14 & 	25.75   \\ 	
                Dense SIFT (VLAD) & 	0.652 & 	28.42 & 	74.98 & 	64.82 & 	37.73 & 	27.77   \\ 	
                  Dense SIFT (FV) & 	0.652 & 	28.42 & 	74.98 & 	64.82 & 	37.73 & 	27.77 \\ 	[4pt]
                                   \emph{CNN-based}&&&&&&  \\
                                   Vgg F & 	0.360 & 	57.22 & 	85.29 & 	80.91 & 	65.26 & 	52.88 \\ 	
                                   Vgg M & 	0.344 & 	58.83 & 	86.55 & 	82.73 & 	67.11 & 54.26 \\ 	
                                   Vgg S & 	0.350 & 	58.34 & 	86.42 & 	82.58 & 	66.81 & 	53.50 \\ 	
                              Vgg M 2048 & 	0.348 & 	58.27 & 	85.63 & 	82.21 & 	67.21 & 	53.82  \\ 	
                              Vgg M 1024 & 	0.358 & 	56.99 & 	85.27 & 	81.79 & 	66.03 & 	52.92   \\ 	
                               Vgg M 128 & 	0.470 & 	44.46 & 	74.04 & 	69.85 & 	53.60 & 	42.39  \\ 	
                                BVLC Ref & 	0.376 & 	55.46 & 	84.84 & 	80.77 & 	63.79 & 	51.34 \\ 	
                            BVLC AlexNet & 	0.388 & 	54.31 & 	84.13 & 	79.37 & 	62.18 & 	50.05  \\ 	
                         Vgg VeryDeep 16 & 	0.365 & 	56.30 & 	84.25 & 	79.88 & 	64.36 & 	52.50  \\ 	
                         Vgg VeryDeep 19 & 	0.369 & 	55.80 & 	83.42 & 	79.08 & 	63.71 & 	51.99  \\
                                          GoogleNet & 	0.293 & 	63.89 & 	97.49 & 	90.85 & 	72.81 & 	58.51 \\
                               ResNet-50 & 	0.305 & 	63.13 & 	98.17 & 	92.54 & 	72.73 & 	57.56 \\
                              ResNet-101 & 	0.301 & 	63.28 & 	98.43 & 	92.47 & 	72.22 & 	57.83 \\
                              ResNet-152 & 	0.308 & 	62.58 & 	98.07 & 	91.87 & 	71.14 & 	57.50 \\ 
                                 NetVLAD & 	0.324 & 	61.03 & 	97.61 & 	90.30 & 	70.53 & 	56.21 \\
                            SatResNet-50 & 	\bf{0.185} & 	\bf{76.55} & 	\bf{98.86} & 	\bf{95.48} & 	\bf{83.79} & 	\bf{69.95} \\
                        
    \bottomrule\\
  \end{tabular}
  \caption{LandUse dataset}
    \end{subtable}
\hspace*{3cm}
 \begin{subtable}{0.5\textwidth}
 \centering
    \begin{tabular}{lcccccccc}
    \toprule
 features 	&	 ANMRR	&	 MAP	&	 P@5		& P@10	&	 P@50	&	 P@100 	\\
 \bottomrule
                                    \emph{Global hand-crafted}&&&&&&  \\
                        Hist. L &	0.732 &	19.73 &	36.82 &	31.14 &	20.49 &	15.85	\\
Hist. H V &	0.713 &	23.04 &	42.77 &	35.80 &	22.36 &	16.56	\\
Hist. RGB &	0.725 &	21.34 &	40.84 &	33.99 &	20.85 &	16.17	\\
Hist. \emph{rgb} &	0.716 &	22.62 &	42.73 &	35.91 &	22.08 &	16.43	\\
Spatial Hist. RGB &	0.742 &	21.55 &	36.92 &	30.25 &	19.77 &	15.14	\\
Co-occ. matr. &	0.825 &	11.57 &	21.00 &	17.64 &	12.72 &	10.85	\\
CEDD &	0.696 &	23.83 &	37.71 &	33.69 &	23.71 &	17.77	\\
DT-CWT L &	0.677 &	23.07 &	35.22 &	30.87 &	23.82 &	19.95\\
DT-CWT &	0.586 &	33.01 &	50.65 &	45.09 &	32.54 &	24.23	\\
Gist RGB &	0.728 &	22.66 &	40.06 &	34.00 &	21.40 &	15.54	\\
Gabor L &	0.692 &	22.38 &	39.52 &	33.86 &	22.84 &	18.81	\\
Gabor RGB &	0.656 &	26.62 &	48.60 &	42.22 &	26.28 &	20.37	\\
Opp. Gabor RGB &	0.647 &	27.77 &	47.54 &	41.31 &	27.85 &	20.52	\\
HoG &	0.733 &	19.81 &	38.75 &	34.10 &	20.96 &	15.38	\\
Granulometry &	0.722 &	21.07 &	38.35 &	32.31 &	20.19 &	17.02	\\
LBP L &	0.701 &	22.12 &	46.43 &	38.80 &	23.09 &	17.50	\\
LBP RGB &	0.672 &	24.70 &	49.35 &	42.85 &	25.62 &	19.03	\\
[4pt]
                                   \emph{Local hand-crafted}&&&&&&  \\
Dense LBP RGB &	0.669 &	24.31 &	50.43 &	42.40 &	25.72 &	19.31	\\
SIFT &	0.570 &	35.64 &	58.55 &	51.11 &	34.25 &	24.84	\\
Dense SIFT &	0.606 &	31.91 &	63.50 &	54.86 &	31.65 &	22.11	\\
Dense SIFT (VLAD) &	0.554 &	37.06 &	70.93 &	61.99 &	36.36 &	24.85	\\
Dense SIFT (FV) &	0.523 &	40.19 &	71.74 &	63.37 &	38.36 &	27.00	\\
[4pt]
                                   \emph{CNN-based}&&&&&&  \\
Vgg F &	0.372 &	54.18 &	72.08 &	70.39 &	53.50 &	34.85	\\
Vgg M &	0.383 &	52.91 &	72.22 &	70.31 &	52.57 &	34.16 	\\
Vgg S &	0.381 &	53.13 &	72.18 &	70.45 &	52.65 &	34.24 	\\
Vgg M 2048 &	0.386 &	52.38 &	71.72 &	69.83 &	52.21 &	34.02	\\
Vgg M 1024 &	0.398 &	51.08 &	70.95 &	69.12 &	50.83 &	33.46 	\\
Vgg M 128 &	0.519 &	38.16 &	60.00 &	56.65 &	39.14 &	27.34 	\\
BVLC Ref &	0.371 &	54.43 &	72.00 &	70.58 &	53.57 &34.88 	\\
BVLC AlexNet &	0.391 &	52.44 &	70.77 &	68.65 &	51.56 &	34.07	\\
Vgg VeryDeep 16 &	0.402 &	50.55 &	71.44 &	69.18 &	50.26 &	33.28	\\
Vgg VeryDeep 19 &	0.419 &	48.61 &	70.01 &	67.17 &	48.58 &	32.48 \\
              GoogleNet & 	0.299 & 	62.12 & 	85.87 & 	81.74 & 	58.14 & 	39.42 \\ 
                               ResNet-50 & 	0.231 & 	70.86 & 	92.42 & 	89.15 & 	65.86 & 	42.27 \\
                              ResNet-101 & 	0.248 & 	68.61 & 	91.88 & 	88.15 & 	63.92 & 	41.50 \\
                              ResNet-152 & 	0.250 & 	68.50 & 	91.76 & 	87.68 & 	63.90 & 	41.27 \\
                                 NetVLAD & 	0.332 & 	59.19 & 	86.25 & 	81.34 & 	55.74 & 	37.08 \\
                            SatResNet-50 & 	\bf{0.027} & 	\bf{96.14} & 	\bf{99.10} & 	\bf{98.74} & 	\bf{93.25} & 	\bf{51.39} \\ 
                        
    \bottomrule\\
  \end{tabular}
   \caption{SceneSat dataset}
        \end{subtable}
 } 
 \caption{Results obtained with the Pseudo RF scheme with the Euclidean distance. The lower is the value of $ANMRR$ and $EQC$ the better is the performance. For the other metrics is the opposite. The best result is reported in bold.}
  \label{tab:land_prf}
\end{table*}

\begin{table}[tb]
 \setlength{\tabcolsep}{2.5pt}
  \def\arraystretch{1.0}%
 \scriptsize
      \resizebox{0.9\textwidth}{!}{
\begin{subtable}{0.5\linewidth}
\centering
  \begin{tabular}{lccccccc}
    \toprule
 features 	&	 ANMRR	&	 MAP	&	 P@5		& P@10	&	 P@50	&	 P@100 	\\
 \bottomrule
                                   \emph{Global hand-crafted}&&&&&&  \\
    Hist. L & 	0.798 & 	15.02 & 	71.40 & 	48.34 & 	21.13 & 	15.24 \\ 	
                              Hist. H V  & 	0.762 & 	18.52 & 	83.90 & 	59.44 & 	26.51 & 	18.39 \\ 	
                               Hist. RGB & 	0.771 & 	17.73 & 	80.50 & 	57.16 & 	25.06 & 	17.65 \\ 	
                        Hist. \emph{rgb} & 	0.782 & 	16.73 & 	79.96 & 	54.60 & 	23.80 & 	16.66\\ 	
                       Spatial Hist. RGB & 	0.789 & 	17.47 & 	82.30 & 	52.48 & 	22.59 & 	16.15 \\ 	
                                              Co-occ. matr. & 	0.853 & 	9.76 & 	38.02 & 	28.78 & 	14.02 & 	10.74   \\ 	
                                    CEDD & 	0.722 & 	22.23 & 	87.97 & 	66.76 & 	32.28 & 	22.04  \\ 	
                                DT-CWT L & 	0.691 & 	23.19 & 	65.54 & 	50.27 & 	29.25 & 	23.32 \\ 	
                                  DT-CWT & 	0.654 & 	27.21 & 	80.47 & 	62.06 & 	35.64 & 	26.85   \\ 	
                                Gist RGB & 	0.767 & 	20.40 & 	84.43 & 	57.66 & 	26.08 & 	18.12  \\ 	
                                 Gabor L & 	0.754 & 	17.99 & 	71.10 & 	51.05 & 	24.59 & 	18.57 \\ 	
                               Gabor RGB & 	0.734 & 	20.11 & 	77.10 & 	56.96 & 	28.26 & 	20.32  \\ 	
                          Opp. Gabor RGB & 	0.732 & 	20.80 & 	80.10 & 	58.71 & 	28.55 & 	20.73  \\ 	
                                     HoG & 	0.733 & 	20.49 & 	78.10 & 	57.23 & 	28.59 & 	20.52 \\ 	
                            Granulometry & 	0.770 & 	17.28 & 	67.09 & 	47.36 & 	22.82 & 	17.10  \\ 	
                                   LBP L & 	0.746 & 	18.91 & 	77.58 & 	57.98 & 	29.05 & 	19.98\\ 	
                                 LBP RGB & 	0.737 & 	20.08 & 	82.13 & 	62.40 & 	30.86 & 	20.80  \\ 	
                         [4pt]
                                   \emph{Local hand-crafted}&&&&&&  \\
                                    Dense LBP RGB & 	0.727 & 	21.74 & 	88.73 & 	66.33 & 	32.21 & 	21.76  \\ 	
                        SIFT & 	0.602 & 	32.89 & 	88.78 & 	67.16 & 	40.66 & 	31.04 \\ 	
                       Dense SIFT & 	0.649 & 	28.38 & 	93.43 & 	75.42 & 	38.82 & 	27.86  \\ 	
               Dense SIFT (VLAD) & 	0.623 & 	31.18 & 	94.79 & 	77.60 & 	41.74 & 	30.13   \\ 	
                  Dense SIFT (FV) & 	0.623 & 	31.18 & 	94.79 & 	77.60 & 	41.74 & 	30.13\\ 	[4pt]
                                   \emph{CNN-based}&&&&&&  \\
                                   Vgg F & 	0.329 & 	60.53 & 	97.39 & 	89.99 & 	69.47 & 	55.64 \\ 	
                                   Vgg M & 	0.316 & 	61.85 & 	97.82 & 	91.13 & 	70.93 & 	56.64   \\ 	
                                   Vgg S & 	0.320 & 	61.45 & 	97.58 & 	90.98 & 	70.77 & 	56.04 \\ 	
                              Vgg M 2048 & 	0.316 & 	61.76 & 	97.97 & 	91.20 & 	71.58 & 	\bf{56.66}  \\ 	
                              Vgg M 1024 & 	0.326 & 	60.55 & 	97.78 & 	90.83 & 	70.48 & 	55.73  \\ 	
                               Vgg M 128 & 	0.422 & 	49.53 & 	95.30 & 	83.89 & 	59.85 & 	46.46  \\ 	
                                BVLC Ref & 	0.347 & 	58.63 & 	97.50 & 	89.95 & 	67.71 & 	53.91  \\ 	
                            BVLC AlexNet & 	0.357 & 	57.63 & 	97.26 & 	88.90 & 	66.53 & 	52.72  \\ 	
                         Vgg VeryDeep 16 & 	0.331 & 	60.05 & 	97.24 & 	89.45 & 	68.98 & 	55.48  \\ 	
                         Vgg VeryDeep 19 & 	0.334 & 	59.56 & 	96.97 & 	88.88 & 	68.39 & 	54.92\\
                                     GoogleNet & 	0.257 & 	67.48 & 	86.59 & 	84.09 & 	62.32 & 	41.49 \\
                               ResNet-50 & 	0.181 & 	77.11 & 	93.53 & 	92.07 & 	71.44 & 	44.65 \\
                              ResNet-101 & 	0.196 & 	75.13 & 	92.72 & 	90.94 & 	69.40 & 	44.03\\
                              ResNet-152 & 	0.203 & 	74.62 & 	92.64 & 	90.63 & 	69.39 & 	43.49 \\ 
                                 NetVLAD & 	0.281 & 	66.36 & 	86.35 & 	83.05 & 	61.34 & 	39.60 \\
                            SatResNet-50 & 	\bf{0.014} & 	\bf{98.05} & 	\bf{99.18} & 	\bf{99.31} & 	\bf{95.79} & 	51.78 \\ 
                          
    \bottomrule\\
  \end{tabular}
  \caption{LandUse dataset}
    \end{subtable}
\hspace*{3cm}
 \begin{subtable}{0.5\textwidth}
 \centering
 \begin{tabular}{lcccccc}
    \toprule
 features 	&	 ANMRR	&	 MAP	&	 P@5		& P@10	&	 P@50	&	 P@100 	\\
 \bottomrule
                 \emph{Global hand-crafted}&&&&&&  \\
          Hist. L & 	0.698 & 	24.02 & 	67.76 & 	47.75 & 	23.30 & 	17.22  \\ 	
                              Hist. H V  & 	0.670 & 	28.17 & 	78.95 & 	55.00 & 	25.93 & 	18.30\\ 	
                               Hist. RGB & 	0.686 & 	25.91 & 	72.36 & 	51.58 & 	24.09 & 	17.77 \\ 	
                        Hist. \emph{rgb} & 	0.675 & 	27.44 & 	76.54 & 	54.10 & 	25.35 & 	18.12   \\ 	
                       Spatial Hist. RGB & 	0.682 & 	28.06 & 	83.62 & 	54.21 & 	24.86 & 	17.71  \\ 		
                            Co-occ. matr. & 	0.803 & 	14.04 & 	41.59 & 	30.37 & 	14.59 & 	11.52 \\ 	
                                    CEDD & 	0.653 & 	29.59 & 	75.30 & 	54.08 & 	27.26 & 	19.60 \\ 	
                                DT-CWT L & 	0.648 & 	26.88 & 	62.55 & 	45.02 & 	26.21 & 	21.13 \\ 	
                                  DT-CWT & 	0.544 & 	37.90 & 	78.35 & 	59.78 & 	36.08 & 	26.13 \\ 	
                                Gist RGB & 	0.660 & 	29.42 & 	83.10 & 	56.85 & 	27.20 & 	18.60 \\ 	
                                 Gabor L & 	0.663 & 	26.39 & 	68.52 & 	48.85 & 	25.20 & 	19.86  \\ 	
                               Gabor RGB & 	0.622 & 	31.02 & 	77.15 & 	57.26 & 	29.28 & 	21.74\\ 	
                          Opp. Gabor RGB & 	0.610 & 	32.29 & 	76.50 & 	57.07 & 	30.94 & 	22.03\\ 	
                                     HoG & 	0.688 & 	24.93 & 	72.06 & 	52.61 & 	24.88 & 	17.15 \\ 	
                            Granulometry & 	0.701 & 	24.55 & 	67.02 & 	46.41 & 	22.07 & 	17.63 \\ 	
                                   LBP L & 	0.655 & 	27.47 & 	79.06 & 	56.44 & 	27.07 & 	19.57\\ 	
                                 LBP RGB & 	0.628 & 	29.79 & 	78.91 & 	58.92 & 	29.37 & 	20.93 \\ 	[4pt]
                                   \emph{Local hand-crafted}&&&&&&  \\
                          Dense LBP RGB & 	0.625 & 	29.82 & 	83.16 & 	59.42 & 	29.55 & 	21.11 \\ 	
                        SIFT  & 	0.508 & 	42.25 & 	89.27 & 	68.54 & 	39.56 & 	27.79  \\ 	
                       Dense SIFT & 	0.554 & 	37.83 & 	90.55 & 	71.12 & 	36.35 & 	24.38   \\ 	
                Dense SIFT (VLAD) & 	0.493 & 	43.48 & 	94.37 & 	77.44 & 	41.69 & 	27.70 \\ 	
                 Dense SIFT (FV) & 	0.493 & 	43.48 & 	94.37 & 	77.44 & 	41.69 & 	27.70  \\ 	[4pt]
                                   \emph{CNN-based}&&&&&&  \\
                                   Vgg F & 	0.330 & 	59.44 & 	92.22 & 	80.82 & 	57.44 & 	36.75 \\ 	
                                   Vgg M & 	0.340 & 	58.30 & 	92.36 & 	81.06 & 	56.41 & 	36.10  \\ 	
                                   Vgg S & 	0.336 & 	58.69 & 	92.42 & 	81.15 & 	56.74 & 	36.34  \\ 	
                              Vgg M 2048 & 	0.339 & 	58.46 & 	92.78 & 	81.52 & 	56.65 & 	36.12 \\ 	
                              Vgg M 1024 & 	0.349 & 	57.21 & 	92.78 & 	80.85 & 	55.31 & 	35.73  \\ 	
                               Vgg M 128 & 	0.448 & 	46.57 & 	90.89 & 	74.01 & 	45.39 & 	30.70 \\ 	
                                BVLC Ref & 	0.327 & 	59.89 & 	92.38 & 81.55 & 	57.63 & 	36.80  \\ 	
                            BVLC AlexNet & 	0.343 & 	58.26 & 	92.72 & 	80.74 & 	55.97 & 	36.16  \\ 	
                         Vgg VeryDeep 16 & 	0.359 & 	55.99 & 	92.12 & 	79.91 & 	54.07 & 	35.24  \\ 	
                         Vgg VeryDeep 19 & 	0.368 & 	54.79 & 	91.84 & 	79.23 & 	53.23 & 	34.76   \\ 	
                                   GoogleNet & 	0.215 & 	72.83 & 	98.29 & 	93.00 & 	66.40 & 	43.43 \\
                               ResNet-50 & 	0.153 & 	80.54 & 	99.76 & 	97.28 & 	74.27 & 	45.92 \\ 
                              ResNet-101 & 	0.168 & 	78.56 & 	99.46 & 	96.14 & 	72.24 & 	45.35 \\
                              ResNet-152 & 	0.173 & 	78.30 & 	99.62 & 	96.43 & 	72.43 & 	44.82\\ 
                                 NetVLAD & 	0.229 & 	72.29 & 	98.87 & 	94.09 & 	66.35 & 	41.97\\ 
                            SatResNet-50 & 	\bf{0.009} & 	\bf{98.69} & 	\bf{100.00} & 	\bf{99.99} & 	\bf{96.46} & 	\bf{51.95} \\ 
    \bottomrule\\
  \end{tabular}
   \caption{SceneSat dataset}
        \end{subtable}
 } 
  \caption{Results obtained with the Manual RF scheme with the Euclidean distance. The lower is the value of $ANMRR$ and $EQC$ the better is the performance. For the other metrics is the opposite. For each row the best result is reported in bold.}
  \label{tab:land_mrf}
\end{table}

\subsection{Feature evaluation using the pseudo-RF retrieval scheme}

In the case of pseudo RF, we used the top $n$ images retrieved after the initial query for re-querying the system. The computational cost of such a system is $n$ times higher than the cost of a basic system.

Results obtained choosing $n=5$ are showed in Table~\ref{tab:land_prf}(a) and ~\ref{tab:land_prf}(b) for the LandUse  and SatScene datasets respectively. It can be noticed that, in both cases, the employment of the pseudo RF scheme gives an improvement with respect to the basic retrieval system whatever is the visual descriptor employed. The CNN-based  and local hand-crafted descriptors that, when used in a basic system, obtained the highest precision at level 5 ($Pr@5$), have the largest improvement of performance.
 
Figures~\ref{fig:increment}(a) and (b) show the difference of MAP between the pseudo RF scheme and the basic retrieval scheme, when the Vgg visual descriptor is employed. The value $n$ ranges from 0 (that corresponds to the basic system) to 10.  It can be noticed that the improvement of performance, when  $n$ is equal to 5, is of about $5 \%$ in the case of LandUse and of about $7 \%$ in the case of SceneSat dataset. 

The second columns of tables~\ref{tab:class_land_alrf} and~\ref{tab:class_rs_alrf} show the best performing visual descriptor for each remote sensing image class. In both cases, LandUse and SceneSat, the best performing visual descriptors are quite the same as in the case of the basic retrieval system.

\subsection{Feature evaluation using the manual-RF retrieval scheme}

In manual RF, we used the first $n$ actually relevant images retrieved after the initial query for re-querying the system. The computational cost of such a system is $n$ times higher than the cost of a basic system. The first five relevant images appear, in the worst case (co-occurrence matrix), within the top 50 images, while in the best case (SatResNet-50),  within the top 6 or 7 images (cfr. table~\ref{tab:land_basic}).

Results obtained choosing $n=5$ are showed in Table~\ref{tab:land_mrf}(a) and ~\ref{tab:land_mrf}(b) for the LandUse  and SatScene datasets respectively. It can be noticed that, in both cases, the employment of the manual RF scheme gives an improvement with respect to both the basic retrieval and the pseudo RF systems. The CNN-based  and local hand-crafted descriptors that, when used in a basic system, obtained the highest precision at level 5 ($Pr@5$), have also in this case the largest improvement of performance.

Figures~\ref{fig:increment}(a) and (b) show the difference between the MAP of the manual RF scheme and the basic retrieval scheme, when the Vgg visual descriptor is employed, The value $n$ ranges from 0 (that corresponds to the basic system) to 10.  It can be noticed that for both datasets the improvement of performance is, when $n$ is equal to 5, of about $9 \%$ in the case of LandUse and of about $14 \%$ in the case of SceneSat. The manual RF scheme, when $n$ is equal to 1, achieves the same performance of the pseudo RF when $n$ is equal to 2.

The third columns of tables~\ref{tab:class_land_alrf} and~\ref{tab:class_rs_alrf} show the best performing visual descriptor for each remote sensing image class. In both cases, LandUse and SceneSat, the best performing visual descriptors are quite the same as in the cases of the basic and pseudo-RF retrieval system.

\subsection{Feature evaluation using the active-learning-based-RF retrieval scheme}

We considered the Active-Learning-based RF scheme as presented by Demir et al.~\cite{Bruzzone}. As suggested by the original authors, we considered the following parameters: 10 RF iterations; an initial training set made of 2 relevant and 3 not relevant images; $p=20$ ambiguous images; $h=5$ diverse images; the histogram intersection as measure of similarity between feature vectors. The histogram intersection distance is defined as follows: $$HI(\mathbf{x},\mathbf{y}) = \sum_{l=1}^{L} min(x_{l},y_{l}),$$ where $\mathbf{x}$ and $\mathbf{y}$ are the feature vectors of two generic images and $L$ is the size of the feature vector.

Results are showed in Table~\ref{tab:land_alrf}(a) and ~\ref{tab:land_alrf}(b) for the LandUse  and SatScene datasets respectively.  Regarding the LandUse dataset, it can be noticed that the employment of  this RF scheme gives an improvement with respect to the other retrieval schemes for all the visual descriptors. In the case of CNN-based descriptors the improvement is of about 20\%. Surprisingly, in the case of SceneSat dataset, the employment of the Active-Learning-based RF scheme gives a performance improvement only in the cases of hand-crafted descriptors and most recent CNN architectures like ResNet or NetVLAD. In the best case, that is NetVLAD, the improvement is of about  80\%. It is very interesting to note that for both datasets, the best performing descriptor is the NetVLAD. This is mostly due to the fact that this feature vector compared with the others extracted from different CNN architecture is less sparse. The degree of sparseness of feature vectors makes the Support Vector Machine, that is employed in the case of Active-Learning-based RF scheme, less or more effective.

The fourth columns of tables~\ref{tab:class_land_alrf} and~\ref{tab:class_rs_alrf} show the best performing visual descriptor for each remote sensing image class. In the case of LandUse dataset, the best performing visual descriptors are the CNN-based descriptors, while in the case of SceneSat dataset, the best performing are the local hand-crafted descriptors apart from a few number of classes.

\begin{table}[tb]
  \setlength{\tabcolsep}{2.5pt}
  \def\arraystretch{1.0}%
 \scriptsize
      \resizebox{0.9\textwidth}{!}{
\begin{subtable}{0.5\linewidth}
\centering
  \begin{tabular}{lccccccc}
    \toprule
 features 	&	 ANMRR	&	 MAP	&	 P@5		& P@10	&	 P@50	&	 P@100 	\\
 \bottomrule
                          \emph{Global hand-crafted}&&&&&&  \\
                                                       Hist. L & 	0.753 & 	19.18 & 	71.46 & 	52.51 & 	25.23 & 	18.82\\
                              Hist. H V  & 	0.688 & 	25.57 & 	87.31 & 	70.88 & 	34.65 & 	24.31 \\                               Hist. RGB & 	0.747 & 	20.12 & 	80.07 & 	61.73 & 	27.77 & 	19.50 \\                        Hist. \emph{rgb} & 	0.712 & 	23.25 & 	83.35 & 	63.96 & 	30.50 & 	22.22 \\                       Spatial Hist. RGB & 	0.695 & 	25.27 & 	84.49 & 	64.50 & 	32.39 & 	23.52 \\                            Coocc. matr. & 	0.851 & 	10.28 & 	32.18 & 	24.41 & 	13.94 & 	10.94 \\                                    CEDD & 	0.719 & 	22.72 & 	85.49 & 	68.70 & 	32.43 & 	21.94\\                                DT-CWT L & 	0.757 & 	18.78 & 	56.10 & 	42.78 & 	24.67 & 	18.81 \\                                  DT-CWT & 	0.698 & 	24.48 & 	77.90 & 	63.52 & 	33.58 & 	23.73 \\                                Gist RGB & 	0.662 & 	27.76 & 	87.06 & 	70.17 & 	37.91 & 	27.11 \\                                 Gabor L & 	0.748 & 	19.19 & 	61.67 & 	44.56 & 	23.95 & 	18.61 \\                               Gabor RGB & 	0.709 & 	23.43 & 	69.60 & 	53.30 & 	30.05 & 	22.74 \\                          Opp. Gabor RGB & 	0.634 & 	29.78 & 	82.34 & 	66.71 & 	39.40 & 	29.42 \\                                     HoG & 	0.681 & 	26.18 & 	84.70 & 	68.10 & 	36.06 & 	25.20 \\                            Granulometry & 	0.833 & 	14.39 & 	54.11 & 	36.23 & 	16.92 & 	12.76 \\
                                   LBP L & 	0.791 & 	16.82 & 	64.78 & 	47.73 & 	24.17 & 	16.49 \\ 
                                 LBP RGB & 	0.793 & 	16.77 & 	69.43 & 	51.66 & 	24.26 & 	16.41\\[4pt]
                \emph{Local hand-crafted}&&&&&&  \\
                          Dense LBP RGB & 	0.726 & 	22.41 & 	85.21 & 	68.24 & 	31.99 & 	21.34 \\                        SIFT & 	0.572 & 	35.40 & 	80.31 & 	65.13 & 	44.08 & 	34.64 \\ 
                       Dense SIFT & 	0.631 & 	32.09 & 	90.27 & 	76.93 & 	41.81 & 	29.82 \\                Dense SIFT (VLAD) & 	0.598 & 	34.39 & 	88.87 & 	76.44 & 	43.70 & 	31.92 \\
                                  Dense SIFT (FV) & 	0.465 & 	48.38 & 	98.58 & 	92.97 & 	61.48 & 	44.70 \\ [4pt]
                   \emph{CNN-based}&&&&&&  \\
                                   Vgg F & 	0.256 & 	69.33 & 	99.70 & 	97.47 & 	79.92 & 	63.90 \\                                   Vgg M & 	0.247 & 	71.09 & 	99.39 & 	98.09 & 	82.98 & 	65.54 \\                                   Vgg S & 	0.260 & 	69.45 & 	99.34 & 	96.42 & 	79.88 & 	63.61 \\                              Vgg M 2048 & 	0.248 & 	70.54 & 	98.51 & 	96.27 & 	80.43 & 	65.05 \\                              Vgg M 1024 & 	0.266 & 	68.22 & 	99.09 & 	96.84 & 	78.75 & 	62.73 \\                               Vgg M 128 & 	0.333 & 	60.72 & 	97.70 & 	94.03 & 	73.60 & 	56.17 \\                                BVLC Ref & 	0.292 & 	66.05 & 	98.70 & 	95.98 & 	76.56 & 	60.72 \\                            BVLC AlexNet & 	0.281 & 	66.98 & 	99.49 & 	96.77 & 	77.36 & 	61.58 \\                         Vgg VeryDeep 16 & 	0.292 & 	66.05 & 	98.70 & 	95.98 & 	76.56 & 	60.72\\
                         Vgg VeryDeep 19 & 	0.292 & 	66.05 & 	98.70 & 	95.98 & 	76.56 & 	60.72  \\
                                                       GoogleNet & 	0.209 & 	75.41 & 	99.76 & 	98.84 & 	87.09 & 	69.36 \\ 
                               ResNet-50 & 	0.332 & 	63.34 & 	99.45 & 	98.01 & 	77.00 & 	56.72 \\ 
                              ResNet-101 & 	0.311 & 	65.42 & 	99.59 & 	98.33 & 	79.15 & 	58.97 \\ 
                              ResNet-152 & 	0.318 & 	64.74 & 	99.68 & 	98.38 & 	78.71 & 	57.81 \\ 
                                 NetVLAD & 	\bf{0.144} & 	\bf{82.43} & 	\bf{99.80} & 	\bf{99.32} & 	\bf{91.31} & 	\bf{76.35} \\ 
                            SatResNet-50 & 	0.232 & 	74.40 & 	99.46 & 	98.89 & 	87.44 & 	67.66 \\ 
                    
    \bottomrule\\
   \end{tabular}
  \caption{LandUse dataset}
    \end{subtable}
\hspace*{3cm}
 \begin{subtable}{0.5\textwidth}
 \centering
  \begin{tabular}{lcccccc}
    \toprule
 features 	&	 ANMRR	&	 MAP	&	 P@5		& P@10	&	 P@50	&	 P@100 	\\
 \bottomrule
                     \emph{Global hand-crafted}&&&&&&  \\
                                    Hist. L & 	0.648 & 	28.93 & 	67.96 & 	52.19 & 	27.10 & 	20.01\\
                              Hist. H V  & 	0.574 & 	36.02 & 	76.44 & 	61.02 & 	34.64 & 	23.78 \\
                               Hist. RGB & 	0.646 & 	29.86 & 	74.03 & 	57.51 & 	28.50 & 	19.43 \\
                        Hist. \emph{rgb} & 	0.587 & 	35.72 & 	76.70 & 	62.33 & 	33.68 & 	22.71 \\
                       Spatial Hist. RGB & 	0.549 & 	38.34 & 	80.78 & 	64.05 & 	36.62 & 	25.10 \\
                            Coocc. matr. & 	0.804 & 	14.72 & 	42.03 & 	31.14 & 	14.93 & 	11.25 \\
                                    CEDD & 	0.673 & 	27.15 & 	69.53 & 	53.37 & 	26.10 & 	18.20 \\
                                DT-CWT L & 	0.688 & 	25.51 & 	55.14 & 	42.37 & 	23.85 & 	18.12 \\
                                  DT-CWT & 	0.571 & 	37.60 & 	71.54 & 	57.35 & 	35.32 & 	23.91 \\
                                Gist RGB & 	0.551 & 	38.50 & 	88.12 & 	70.45 & 	36.75 & 	24.55 \\
                                 Gabor L & 	0.672 & 	26.71 & 	65.65 & 	46.94 & 	24.21 & 	19.31 \\
                               Gabor RGB & 	0.612 & 	32.21 & 	70.51 & 	55.46 & 	30.40 & 	22.13 \\
                          Opp. Gabor RGB & 	0.473 & 	46.13 & 	86.77 & 	74.32 & 	44.11 & 	29.00\\
                                     HoG & 	0.699 & 	24.17 & 	73.53 & 	53.95 & 	23.91 & 	16.37 \\
                            Granulometry & 	0.794 & 	18.23 & 	53.31 & 	37.13 & 	15.88 & 	11.59 \\
                                   LBP L & 	0.669 & 	26.62 & 	72.44 & 	54.86 & 	26.68 & 	18.24 \\
                                 LBP RGB & 	0.667 & 	27.26 & 	65.65 & 	49.47 & 	26.07 & 	19.14 \\[4pt]
                          \emph{Local hand-crafted}&&&&&&  \\
                          Dense LBP RGB & 	0.534 & 	39.84 & 	84.20 & 	68.96 & 	37.18 & 	26.25 \\ 
                       SIFT  & 	0.473 & 	46.13 & 	89.61 & 	76.34 & 	43.64 & 	29.25 \\
                       Dense SIFT & 	0.455 & 	48.71 & 	92.30 & 	80.85 & 	45.83 & 	29.47 \\                Dense SIFT (VLAD) & 	0.396 & 	54.98 & 	91.98 & 	84.73 & 	50.72 & 	33.14 \\
               Dense SIFT (FV) & 	0.301 & 	65.81 & 	97.31 & 	94.01 & 	60.48 & 	37.57 \\ [4pt]
                                   \emph{CNN-based}&&&&&&  \\
                                Vgg F &      0.520 &         44.87 &         83.66 &         68.39 &         41.36 &         25.67 \\
                                   Vgg M &      0.503 &         46.40 &         84.82 &         70.47 &         42.82 &         26.35 \\
                               Vgg M 128 &      0.558 &         39.27 &         81.57 &         67.82 &         37.05 &         23.81 \\
                                   Vgg S &      0.479 &         48.61 &         84.90 &         72.35 &         44.81 &         27.90 \\ 
                              Vgg M 2048 &      0.467 &         48.85 &         84.60 &         72.33 &         46.05 &         28.59 \\ 
                              Vgg M 1024 &      0.468 &         48.55 &         85.57 &         72.21 &         45.58 &         28.61 \\ 
                                BVLC Ref &      0.522 &         44.41 &         82.75 &         68.66 &         41.30 &         25.38\\ 
                            BVLC AlexNet &      0.506 &         46.57 &         82.67 &         69.22 &         42.32 &         26.41 \\
                         Vgg VeryDeep 16 &      0.505 &         46.77 &         86.97 &         72.49 &         42.40 &         26.27 \\ 
                         Vgg VeryDeep 19 &      0.498 &         46.83 &         82.53 &         70.32 &         43.55 &         26.59 \\
                            GoogleNet & 	0.112 & 	86.56 & 	99.84 & 	99.34 & 	81.44 & 	46.96 \\
                               ResNet-50 & 	0.152 & 	82.46 & 	99.96 & 	99.82 & 	76.96 & 	44.73 \\ 
                              ResNet-101 & 	0.162 & 	81.52 & 	99.96 & 	99.64 & 	75.86 & 	44.16 \\
                              ResNet-152 & 	0.161 & 	81.46 & 	99.94 & 	99.62 & 	75.80 & 	44.26 \\
                                 NetVLAD & 	\bf{0.048} & 	93.89 & \bf{	100.00} & 	99.94 & 	89.11 & 	\bf{50.28} \\ 
                            SatResNet-50 & 	0.054 & \bf{	94.20} & 	99.98 & 	\bf{99.98} & 	\bf{91.42} & 	49.29 \\
                                   \bottomrule\\
   \end{tabular}
   \caption{SceneSat dataset}
        \end{subtable}
 } 
 \caption{Results obtained with the Active-Learning-based RF scheme. The lower is the value of $ANMRR$ and $EQC$ the better is the performance. For the other metrics is the opposite. For each row the best result is reported in bold.}
  \label{tab:land_alrf}
\end{table}

\begin{table}[h!]
\renewcommand{\arraystretch}{1.7}
 \tiny
 \centering
       \resizebox{0.7\textwidth}{!}{
  \centering
  \begin{tabular}{llcccc}
    categories&image&basic IR&pseudo RF&manual RF& act. learn. RF\\
    \toprule
\multirow{2}{*}{agricultural}& 	\multirow{2}{*}{\includegraphics[width=0.06\textwidth]{agricultural00}}& 	\cellcolor{blue!50}Vgg M& 	\cellcolor{blue!50}ResNet-101& 	\cellcolor{blue!50}Vgg VeryDeep 19& 	\cellcolor{yellow!50} NetVLAD\\
& 	& 	(0.092)& 	(0.065)& 	(0.048)& 	(0.011)\\[3pt]
\hline
\multirow{2}{*}{airplane}& 	\multirow{2}{*}{\includegraphics[width=0.06\textwidth]{airplane00}}& 	\cellcolor{orange!50}SatResNet-50& 	\cellcolor{orange!50}SatResNet-50& 	\cellcolor{orange!50}SatResNet-50& \cellcolor{yellow!50}	NetVLAD\\
& 	& 	(0.148)& 	(0.103)& 	(0.084)& 	(0.033)\\[3pt]
\hline
\multirow{2}{*}{baseballdiamond}& 	\multirow{2}{*}{\includegraphics[width=0.06\textwidth]{baseballdiamond00}}& 	\cellcolor{orange!50}SatResNet-50& 	\cellcolor{orange!50}SatResNet-50& 	\cellcolor{orange!50}SatResNet-50& 	\cellcolor{blue!50}BVLC Ref\\
& 	& 	(0.109)& 	(0.060)& 	(0.059)& 	(0.076)\\[3pt]
\hline
\multirow{2}{*}{beach}& 	\multirow{2}{*}{\includegraphics[width=0.06\textwidth]{beach00}}& 	\cellcolor{orange!50}SatResNet-50& 	\cellcolor{orange!50}SatResNet-50& 	\cellcolor{orange!50}SatResNet-50& 	\cellcolor{blue!50}Vgg F\\
& 	& 	(0.031)& 	(0.021)& 	(0.021)& 	(0.006)\\[3pt]
\hline
\multirow{2}{*}{buildings}& 	\multirow{2}{*}{\includegraphics[width=0.06\textwidth]{buildings00}}& 	\cellcolor{orange!50}SatResNet-50& 	\cellcolor{orange!50}SatResNet-50& 	\cellcolor{orange!50}SatResNet-50& 	\cellcolor{orange!50}SatResNet-50\\
& 	& 	(0.412)& 	(0.368)& 	(0.302)& 	(0.271)\\[3pt]
\hline
\multirow{2}{*}{chaparral}& 	\multirow{2}{*}{\includegraphics[width=0.06\textwidth]{chaparral00}}& 	\cellcolor{yellow!50} NetVLAD& 	\cellcolor{yellow!50} NetVLAD& 	\cellcolor{yellow!50} NetVLAD& 	\cellcolor{yellow!50} NetVLAD\\
& 	& 	(0.007)& 	(0.003)& 	(0.003)& 	(0.001)\\[3pt]
\hline
\multirow{2}{*}{denseresidential}& 	\multirow{2}{*}{\includegraphics[width=0.06\textwidth]{denseresidential00}}& 	\cellcolor{orange!50}SatResNet-50& 	\cellcolor{orange!50}SatResNet-50& 	\cellcolor{orange!50}SatResNet-50& 	\cellcolor{yellow!50} NetVLAD\\
& 	& 	(0.564)& 	(0.561)& 	(0.521)& 	(0.444)\\[3pt]
\hline
\multirow{2}{*}{forest}& 	\multirow{2}{*}{\includegraphics[width=0.06\textwidth]{forest00}}& 	\cellcolor{orange!50}SatResNet-50& 	\cellcolor{blue!50}Vgg M& 	\cellcolor{blue!50}Vgg M& \cellcolor{yellow!50} 	NetVLAD\\
& 	& 	(0.035)& 	(0.023)& 	(0.021)& 	(0.008)\\[3pt]
\hline
\multirow{2}{*}{freeway}& 	\multirow{2}{*}{\includegraphics[width=0.06\textwidth]{freeway00}}& 	\cellcolor{orange!50}SatResNet-50& \cellcolor{orange!50}	SatResNet-50& 	\cellcolor{orange!50}SatResNet-50& 	\cellcolor{yellow!50} NetVLAD\\
& 	& 	(0.280)& 	(0.260)& 	(0.256)& 	(0.142)\\[3pt]
\hline
\multirow{2}{*}{golfcourse}& 	\multirow{2}{*}{\includegraphics[width=0.06\textwidth]{golfcourse00}}& 	\cellcolor{orange!50}SatResNet-50& 	\cellcolor{orange!50}SatResNet-50& 	\cellcolor{orange!50}SatResNet-50& 	\cellcolor{yellow!50} NetVLAD\\
& 	& 	(0.232)& 	(0.181)& 	(0.169)& 	(0.092)\\[3pt]
\hline
\multirow{2}{*}{harbor}& 	\multirow{2}{*}{\includegraphics[width=0.06\textwidth]{harbor00}}& 	\cellcolor{yellow!50} NetVLAD& 	\cellcolor{yellow!50} NetVLAD& 	\cellcolor{yellow!50} NetVLAD& 	\cellcolor{yellow!50} NetVLAD\\
& 	& 	(0.069)& 	(0.051)& 	(0.051)& 	(0.001)\\[3pt]
\hline
\multirow{2}{*}{intersection}& 	\multirow{2}{*}{\includegraphics[width=0.06\textwidth]{intersection00}}& 	\cellcolor{orange!50}SatResNet-50& 	\cellcolor{orange!50}SatResNet-50& 	\cellcolor{orange!50}SatResNet-50& 	\cellcolor{orange!50}SatResNet-50\\
& 	& 	(0.356)& 	(0.289)& 	(0.262)& 	(0.212)\\[3pt]
\hline
\multirow{2}{*}{mediumresidential}& 	\multirow{2}{*}{\includegraphics[width=0.06\textwidth]{mediumresidential00}}& 	\cellcolor{orange!50}SatResNet-50& 	\cellcolor{orange!50}SatResNet-50& \cellcolor{orange!50}	SatResNet-50& 	\cellcolor{yellow!50} NetVLAD\\
& 	& 	(0.390)& 	(0.360)& 	(0.337)& 	(0.292)\\[3pt]
\hline
\multirow{2}{*}{mobilehomepark}& 	\multirow{2}{*}{\includegraphics[width=0.06\textwidth]{mobilehomepark00}}& 	\cellcolor{orange!50}SatResNet-50& 	\cellcolor{orange!50}SatResNet-50& 	\cellcolor{orange!50}SatResNet-50& 	\cellcolor{yellow!50} NetVLAD\\
& 	& 	(0.329)& 	(0.304)& 	(0.278)& 	(0.155)\\[3pt]
\hline
\multirow{2}{*}{overpass}& 	\multirow{2}{*}{\includegraphics[width=0.06\textwidth]{overpass00}}& 	\cellcolor{orange!50}SatResNet-50& \cellcolor{orange!50}	SatResNet-50& 	\cellcolor{orange!50}SatResNet-50& 	\cellcolor{yellow!50} NetVLAD\\
& 	& 	(0.190)& 	(0.152)& 	(0.130)& 	(0.150)\\[3pt]
\hline
\multirow{2}{*}{parkinglot}& 	\multirow{2}{*}{\includegraphics[width=0.06\textwidth]{parkinglot00}}& 	\cellcolor{orange!50}SatResNet-50& 	\cellcolor{orange!50}SatResNet-50& 	\cellcolor{orange!50}SatResNet-50& 	\cellcolor{blue!50}Vgg F\\
& 	& 	(0.002)& 	(0.001)& 	(0.001)& 	(0.006)\\[3pt]
\hline
\multirow{2}{*}{river}& 	\multirow{2}{*}{\includegraphics[width=0.06\textwidth]{river00}}& 	\cellcolor{orange!50}SatResNet-50& \cellcolor{orange!50}	SatResNet-50& 	\cellcolor{orange!50}SatResNet-50& 	\cellcolor{yellow!50} NetVLAD\\
& 	& 	(0.365)& 	(0.317)& 	(0.283)& 	(0.196)\\[3pt]
\hline
\multirow{2}{*}{runway}& 	\multirow{2}{*}{\includegraphics[width=0.06\textwidth]{runway00}}& 	\cellcolor{blue!50}GoogleNet& 	\cellcolor{blue!50}GoogleNet& 	\cellcolor{blue!50}GoogleNet& 	\cellcolor{yellow!50} NetVLAD\\
& 	& 	(0.256)& 	(0.194)& 	(0.191)& 	(0.061)\\[3pt]
\hline
\multirow{2}{*}{sparseresidential}& 	\multirow{2}{*}{\includegraphics[width=0.06\textwidth]{sparseresidential00}}& 	\cellcolor{orange!50}SatResNet-50& 	\cellcolor{orange!50}SatResNet-50& 	\cellcolor{orange!50}SatResNet-50& 	\cellcolor{blue!50}GoogleNet\\
& 	& 	(0.194)& 	(0.127)& 	(0.105)& 	(0.123)\\[3pt]
\hline
\multirow{2}{*}{storagetanks}& 	\multirow{2}{*}{\includegraphics[width=0.06\textwidth]{storagetanks00}}& 	\cellcolor{orange!50}SatResNet-50& 	\cellcolor{orange!50}SatResNet-50& 	\cellcolor{orange!50}SatResNet-50& 	\cellcolor{yellow!50} NetVLAD\\
& 	& 	(0.363)& 	(0.331)& 	(0.278)& 	(0.176)\\[3pt]
\hline
\multirow{2}{*}{tenniscourt}& 	\multirow{2}{*}{\includegraphics[width=0.06\textwidth]{tenniscourt00}}& 	\cellcolor{orange!50}SatResNet-50& 	\cellcolor{orange!50}SatResNet-50& 	\cellcolor{orange!50}SatResNet-50& 	\cellcolor{blue!50}GoogleNet\\
& 	& 	(0.402)& 	(0.360)& 	(0.275)& 	(0.216)\\[3pt]
\bottomrule
  \end{tabular}
  }
  \caption{LandUse dataset: ANMRR evaluation across retrieval schemes and RS image classes . For each class and retrieval scheme is reported the best visual descriptor. Orange color stands for fine-tuned CNN-based descriptors, blue color stands for pre-trained CNN-based descriptors, yellow color stands for NetVLAD-based descriptors while  cyan color stands for global hand-crafted descriptors.}
  \label{tab:class_land_alrf}
\end{table}

\begin{table}[h!]
\renewcommand{\arraystretch}{1.7}
 \tiny
 \centering
      \resizebox{0.7\textwidth}{!}{
  \centering
  \begin{tabular}{llcccc}
    categories&image&basic IR&pseudo RF&manual RF& act. learn. RF\\
    \toprule
\multirow{2}{*}{Airport}& 	\multirow{2}{*}{\includegraphics[width=0.06\textwidth]{airport_01}}& 	\cellcolor{orange!50}SatResNet-50& 	\cellcolor{orange!50}SatResNet-50& 	\cellcolor{orange!50}SatResNet-50& 	\cellcolor{yellow!50} NetVLAD\\
& 	& 	(0.049)& 	(0.015)& 	(0.008)& 	(0.015)\\[3pt]
\hline
\multirow{2}{*}{Beach}& 	\multirow{2}{*}{\includegraphics[width=0.06\textwidth]{beach-01}}& 	\cellcolor{orange!50}SatResNet-50& 	\cellcolor{orange!50}SatResNet-50& 	\cellcolor{blue!50}Vgg M& 	\cellcolor{orange!50}SatResNet-50\\
& 	& 	(0.046)& 	(0.040)& 	(0.037)& 	(0.004)\\[3pt]
\hline
\multirow{2}{*}{Bridge}& 	\multirow{2}{*}{\includegraphics[width=0.06\textwidth]{bridge_01}}& 	\cellcolor{orange!50}SatResNet-50& \cellcolor{orange!50}	SatResNet-50& 	\cellcolor{orange!50}SatResNet-50& 	\cellcolor{blue!50}GoogleNet\\
& 	& 	(0.021)& 	(0.001)& 	(0.001)& 	(0.048)\\[3pt]
\hline
\multirow{2}{*}{Commercial}& 	\multirow{2}{*}{\includegraphics[width=0.06\textwidth]{commercial_01}}& 	\cellcolor{orange!50}SatResNet-50& 	\cellcolor{orange!50}SatResNet-50& 	\cellcolor{orange!50}SatResNet-50& 	\cellcolor{orange!50}SatResNet-50\\
& 	& 	(0.027)& 	(0.008)& 	(0.005)& 	(0.084)\\[3pt]
\hline
\multirow{2}{*}{Desert}& 	\multirow{2}{*}{\includegraphics[width=0.06\textwidth]{Desert_01}}& 	\cellcolor{orange!50}SatResNet-50& 	\cellcolor{orange!50}SatResNet-50& 	\cellcolor{orange!50}SatResNet-50& 	\cellcolor{cyan!50}Opp. Gabor RGB\\
& 	& 	(0.005)& 	(0.003)& 	(0.003)& 	(0.004)\\[3pt]
\hline
\multirow{2}{*}{Farmland}& 	\multirow{2}{*}{\includegraphics[width=0.06\textwidth]{Farmland-01}}& 	\cellcolor{orange!50}SatResNet-50& 	\cellcolor{orange!50}SatResNet-50& 	\cellcolor{orange!50}SatResNet-50& 	\cellcolor{yellow!50} NetVLAD\\
& 	& 	(0.054)& 	(0.034)& 	(0.013)& 	(0.027)\\[3pt]
\hline
\multirow{2}{*}{footballField}& 	\multirow{2}{*}{\includegraphics[width=0.06\textwidth]{footballField_01}}& 	\cellcolor{orange!50}SatResNet-50& 	\cellcolor{orange!50}SatResNet-50& 	\cellcolor{orange!50}SatResNet-50& 	\cellcolor{orange!50}SatResNet-50\\
& 	& 	(0.001)& 	(0.001)& 	(0.001)& 	(0.001)\\[3pt]
\hline
\multirow{2}{*}{Forest}& 	\multirow{2}{*}{\includegraphics[width=0.06\textwidth]{forest_01}}& 	\cellcolor{orange!50}SatResNet-50& \cellcolor{orange!50}	SatResNet-50& 	\cellcolor{orange!50}SatResNet-50& 	\cellcolor{yellow!50} NetVLAD\\
& 	& 	(0.017)& 	(0.014)& 	(0.005)& 	(0.033)\\[3pt]
\hline
\multirow{2}{*}{Industrial}& 	\multirow{2}{*}{\includegraphics[width=0.06\textwidth]{industrial_01}}& 	\cellcolor{orange!50}SatResNet-50& \cellcolor{orange!50}	SatResNet-50& 	\cellcolor{orange!50}SatResNet-50& 	\cellcolor{orange!50}SatResNet-50\\
& 	& 	(0.073)& 	(0.050)& 	(0.028)& 	(0.145)\\[3pt]
\hline
\multirow{2}{*}{Meadow}& 	\multirow{2}{*}{\includegraphics[width=0.06\textwidth]{meadow_01}}& 	\cellcolor{orange!50}SatResNet-50& 	\cellcolor{orange!50}SatResNet-50& 	\cellcolor{orange!50}SatResNet-50& \cellcolor{yellow!50}	NetVLAD\\
& 	& 	(0.020)& 	(0.008)& 	(0.006)& 	(0.035)\\[3pt]
\hline
\multirow{2}{*}{Mountain}& 	\multirow{2}{*}{\includegraphics[width=0.06\textwidth]{Mountain_01}}& 	\cellcolor{orange!50}SatResNet-50& 	\cellcolor{orange!50}SatResNet-50& 	\cellcolor{orange!50}SatResNet-50& 	\cellcolor{yellow!50} NetVLAD\\
& 	& 	(0.014)& 	(0.005)& 	(0.005)& 	(0.003)\\[3pt]
\hline
\multirow{2}{*}{Park}& 	\multirow{2}{*}{\includegraphics[width=0.06\textwidth]{Park_01}}& 	\cellcolor{orange!50}SatResNet-50& 	\cellcolor{orange!50}SatResNet-50& 	\cellcolor{orange!50}SatResNet-50& 	\cellcolor{orange!50}SatResNet-50\\
& 	& 	(0.030)& 	(0.014)& 	(0.008)& 	(0.012)\\[3pt]
\hline
\multirow{2}{*}{Parking}& 	\multirow{2}{*}{\includegraphics[width=0.06\textwidth]{parking_01}}& 	\cellcolor{orange!50}SatResNet-50& 	\cellcolor{orange!50}SatResNet-50& 	\cellcolor{orange!50}SatResNet-50& 	\cellcolor{yellow!50} NetVLAD\\
& 	& 	(0.003)& 	(0.001)& 	(0.001)& 	(0.010)\\[3pt]
\hline
\multirow{2}{*}{Pond}& 	\multirow{2}{*}{\includegraphics[width=0.06\textwidth]{pond_01}}& 	\cellcolor{orange!50}SatResNet-50& 	\cellcolor{orange!50}SatResNet-50& 	\cellcolor{orange!50}SatResNet-50& \cellcolor{orange!50}	SatResNet-50\\
& 	& 	(0.014)& 	(0.004)& 	(0.003)& 	(0.030)\\[3pt]
\hline
\multirow{2}{*}{Port}& 	\multirow{2}{*}{\includegraphics[width=0.06\textwidth]{port_01}}& 	\cellcolor{orange!50}SatResNet-50& 	\cellcolor{orange!50}SatResNet-50& 	\cellcolor{orange!50}SatResNet-50& 	\cellcolor{yellow!50} NetVLAD\\
& 	& 	(0.048)& 	(0.016)& 	(0.010)& 	(0.058)\\[3pt]
\hline
\multirow{2}{*}{railwayStation}& 	\multirow{2}{*}{\includegraphics[width=0.06\textwidth]{railwayStation_01}}& 	\cellcolor{orange!50}SatResNet-50& 	\cellcolor{orange!50}SatResNet-50& 	\cellcolor{orange!50}SatResNet-50& 	\cellcolor{orange!50}SatResNet-50\\
& 	& 	(0.016)& 	(0.021)& 	(0.004)& 	(0.031)\\[3pt]
\hline
\multirow{2}{*}{Residential}& 	\multirow{2}{*}{\includegraphics[width=0.06\textwidth]{residential_01}}& 	\cellcolor{orange!50}SatResNet-50& 	\cellcolor{orange!50}SatResNet-50& 	\cellcolor{orange!50}SatResNet-50& 	\cellcolor{yellow!50} NetVLAD\\
& 	& 	(0.075)& 	(0.034)& 	(0.031)& 	(0.033)\\[3pt]
\hline
\multirow{2}{*}{River}& 	\multirow{2}{*}{\includegraphics[width=0.06\textwidth]{river_01}}& 	\cellcolor{orange!50}SatResNet-50& 	\cellcolor{orange!50}SatResNet-50& 	\cellcolor{orange!50}SatResNet-50& 	\cellcolor{orange!50}SatResNet-50\\
& 	& 	(0.001)& 	(0.001)& 	(0.001)& 	(0.005)\\[3pt]
\hline
\multirow{2}{*}{Viaduct}& 	\multirow{2}{*}{\includegraphics[width=0.06\textwidth]{viaduct_01}}& 	\cellcolor{orange!50}SatResNet-50&\cellcolor{orange!50} 	SatResNet-50& 	\cellcolor{orange!50}SatResNet-50& 	\cellcolor{yellow!50} NetVLAD\\
& 	& 	(0.001)& 	(0.001)& 	(0.001)& 	(0.005)\\[3pt]
\bottomrule
 \end{tabular}
 }
  \caption{LandUse dataset: ANMRR evaluation across retrieval schemes and RS image classes . For each class and retrieval scheme is reported the best visual descriptor. Orange color stands for fine-tuned CNN-based descriptors, blue color stands for pre-trained CNN-based descriptors, yellow color stands for NetVLAD-based descriptors while  cyan color stands for global hand-crafted descriptors.}
  \label{tab:class_rs_alrf}
\end{table}

\begin{table}[tb]
 \tiny
 \centering
       \resizebox{0.9\textwidth}{!}{
  \centering
  \begin{tabular}{l|cc|cc|cc|cc|c}
    \toprule
     	 	& \multicolumn{2}{c}{basic IR}	&	 \multicolumn{2}{c}{pseudo RF}	&	 \multicolumn{2}{c}{manual RF}	& \multicolumn{2}{c|}{act. learn. RF}	&	 Overall	 	\\[2pt]
   \cmidrule(lr){2-3}\cmidrule(lr){4-5}\cmidrule(lr){6-7}\cmidrule(lr){8-9}
        features  	&	\emph{ANMRR}&\emph{EQC}	&	\emph{ANMRR}&\emph{EQC}	& \emph{ANMRR}&\emph{EQC}&	  \emph{ANMRR}&\emph{EQC}	&\emph{avg rank}		\\
 \bottomrule
   SatResNet-50 & 	\bf{0.133} & 	409 & 	\bf{0.110} & 	2045 & 	\bf{0.097} & 	2045 & 	\bf{0.143} & 	8180 & 	5.08 \\
                               GoogleNet & 	0.329 & 	204 & 	0.290 & 	1020 & 	0.254 & 	1020 & 	0.160 & 	4080 & 	5.49 \\
                               ResNet-50 & 	0.294 & 	409 & 	0.255 & 	2045 & 	0.229 & 	2045 & 	0.241 & 	8180 & 	5.72 \\
                              ResNet-101 & 	0.302 & 	409 & 	0.261 & 	2045 & 	0.234 & 	2045 & 	0.236 & 	8180 & 	5.92 \\
                              ResNet-152 & 	0.305 & 	409 & 	0.267 & 	2045 & 	0.240 & 	2045 & 	0.238 & 	8180 & 	6.59 \\
                              Vgg M 2048 & 	0.410 & 	409 & 	0.367 & 	2045 & 	0.327 & 	2045 & 	0.375 & 	8180 & 	8.87 \\
                                 NetVLAD & 	0.369 & 	819 & 	0.326 & 	4095 & 	0.277 & 	4095 & 	0.096 & 	16380 & 	9.02 \\
                              Vgg M 1024 & 	0.422 & 	204 & 	0.378 & 	1020 & 	0.337 & 	1020 & 	0.380 & 	4080 & 	9.19 \\
                                   Vgg M & 	0.398 & 	819 & 	0.363 & 	4095 & 	0.328 & 	4095 & 	0.386 & 	16380 & 	9.45 \\
                                   Vgg S & 	0.398 & 	819 & 	0.365 & 	4095 & 	0.328 & 	4095 & 	0.371 & 	16380 & 	9.55 \\
                                   Vgg F & 	0.397 & 	819 & 	0.366 & 	4095 & 	0.329 & 	4095 & 	0.398 & 	16380 & 	9.90 \\
                               Vgg M 128 & 	0.525 & 	25 & 	0.494 & 	125 & 	0.435 & 	125 & 	0.463 & 	500 & 	10.81 \\
                                BVLC Ref & 	0.404 & 	819 & 	0.374 & 	4095 & 	0.337 & 	4095 & 	0.409 & 	16380 & 	11.18 \\
                                  DT-CWT & 	0.628 & 	4 & 	0.632 & 	20 & 	0.599 & 	20 & 	0.634 & 	80 & 	11.22 \\
                         Vgg VeryDeep 19 & 	0.427 & 	819 & 	0.394 & 	4095 & 	0.351 & 	4095 & 	0.307 & 	16380 & 	11.81 \\
                         Vgg VeryDeep 16 & 	0.416 & 	819 & 	0.383 & 	4095 & 	0.345 & 	4095 & 	0.403 & 	16380 & 	12.02 \\
                            BVLC AlexNet & 	0.416 & 	819 & 	0.389 & 	4095 & 	0.350 & 	4095 & 	0.403 & 	16380 & 	12.03 \\
                                    SIFT & 	0.597 & 	204 & 	0.609 & 	1020 & 	0.555 & 	1020 & 	0.522 & 	4080 & 	12.45 \\
                              Dense SIFT & 	0.638 & 	204 & 	0.641 & 	1020 & 	0.602 & 	1020 & 	0.543 & 	4080 & 	12.61 \\
                          Opp. Gabor RGB & 	0.692 & 	52 & 	0.698 & 	260 & 	0.671 & 	260 & 	0.553 & 	1040 & 	13.56 \\
                                DT-CWT L & 	0.690 & 	1 & 	0.693 & 	5 & 	0.670 & 	5 & 	0.655 & 	20 & 	13.74 \\
                         Dense SIFT (FV) & 	0.579 & 	8192 & 	0.582 & 	40960 & 	0.535 & 	40960 & 	0.498 & 	163840 & 	13.82 \\
                       Dense SIFT (VLAD) & 	0.601 & 	5120 & 	0.603 & 	25600 & 	0.559 & 	25600 & 	0.384 & 	102400 & 	13.93 \\
                           Dense LBP RGB & 	0.702 & 	204 & 	0.708 & 	1020 & 	0.676 & 	1020 & 	0.631 & 	4080 & 	14.10 \\
                               Gabor RGB & 	0.699 & 	19 & 	0.703 & 	95 & 	0.678 & 	95 & 	0.659 & 	380 & 	14.27 \\
                                 LBP RGB & 	0.708 & 	10 & 	0.713 & 	50 & 	0.683 & 	50 & 	0.730 & 	200 & 	14.82 \\
                                    CEDD & 	0.710 & 	28 & 	0.718 & 	140 & 	0.687 & 	140 & 	0.696 & 	560 & 	14.95 \\
                              Hist. H V  & 	0.741 & 	102 & 	0.747 & 	510 & 	0.715 & 	510 & 	0.630 & 	2040 & 	15.55 \\
                                 Gabor L & 	0.726 & 	6 & 	0.731 & 	30 & 	0.709 & 	30 & 	0.709 & 	120 & 	15.76 \\
                                   LBP L & 	0.725 & 	48 & 	0.732 & 	240 & 	0.701 & 	240 & 	0.729 & 	960 & 	16.01 \\
                                Gist RGB & 	0.743 & 	102 & 	0.765 & 	510 & 	0.713 & 	510 & 	0.606 & 	2040 & 	16.53 \\
                                     HoG & 	0.737 & 	16 & 	0.744 & 	80 & 	0.710 & 	80 & 	0.688 & 	320 & 	16.91 \\
                        Hist. \emph{rgb} & 	0.750 & 	153 & 	0.759 & 	765 & 	0.728 & 	765 & 	0.648 & 	3060 & 	17.18 \\
                            Granulometry & 	0.748 & 	15 & 	0.752 & 	75 & 	0.735 & 	75 & 	0.813 & 	300 & 	17.39 \\
                                 Hist. L & 	0.772 & 	51 & 	0.776 & 	255 & 	0.748 & 	255 & 	0.661 & 	1020 & 	18.11 \\
                               Hist. RGB & 	0.754 & 	153 & 	0.757 & 	765 & 	0.728 & 	765 & 	0.695 & 	3060 & 	18.35 \\
                       Spatial Hist. RGB & 	0.763 & 	307 & 	0.783 & 	1535 & 	0.735 & 	1535 & 	0.621 & 	6140 & 	18.67 \\
                            Coocc. matr. & 	0.841 & 	1 & 	0.844 & 	5 & 	0.828 & 	5 & 	0.827 & 	20 & 	19.08 \\
                            \bottomrule \\
  \end{tabular}
  }
  \caption{Average rank across datasets of each visual descriptor performance. The list of visual descriptors reported in the table is ordered by the average rank (last column of the table) that is obtained by averaging the ranks achieved by  each visual descriptor across datasets, retrieval schemes  and measures: $ANMRR$, $MAP$, $Pr$ at 5,10,50,100 levels, and EQC. For sake of completeness, for each retrieval scheme, the table shows the average ANMRR across datasets and the EQC for each visual descriptor. For each retrieval scheme, the best  average ANMRR performance is reported in bold.}
  \label{tab:summary}
\end{table}

\subsection{Average rank of visual descriptors across RS datasets}
In table~\ref{tab:summary} we show the average rank of all the visual descriptors evaluated. The average rank is represented in the last column and obtained by averaging the ranks achieved by  each visual descriptor across datasets, retrieval schemes  and measures: $ANMRR$, $MAP$, $Pr$ at 5,10,50,100 levels, and EQC. For sake of completeness, for each retrieval scheme, we displayed the average ANMRR across datasets and the EQC for each visual descriptor. From this table is quite clear that across datasets, the best performing visual descriptors are the CNN-based ones. The first 13 positions out of 38 are occupied by CNN-based descriptors. The global hand-crafted descriptor DT-CWT is at 14th position mostly because of the length of the vector that is very short. After some other CNN-based descriptors, we find the local hand-crafted descriptors that despite their good performance, they are penalized by the size of the vector of feature that is very long, in the case of Dense SIFT (FV) is 40960 that is 2048 times higher than the size of DT-CWT.

Looking at the EQC columns of each retrieval schemes of table~\ref{tab:summary}, it is quite evident that the use of  Active-Learning-based RF is not always convenient. For instance, in the case of the top 5 visual descriptors of the table, the Active-Learning-based RF achieves globally worse performance than pseudo-RF with a much more higher EQC. This is not true in all other cases, where the performance achieved with the Active-Learning-based RF is better than pseudo-RF.

Notwithstanding this, the employment of techniques to speed-up the nearest image search process makes the AL-RF scheme not as computationally expensive as argued in the previous paragraph. Large amount of data and high dimensional feature vector, makes the nearest image search process very slow. The main bottleneck of the search is the access to the memory. The employment of a compact representation of the feature vectors, such as hash~\cite{zhao2015deep} or polysemous codes~\cite{douze2016polysemous}, is likely to offer a better efficiency than the use of full vectors thus accelerating the image search process.  Readers who would wish to deepen the subject can refer to the following papers~\cite{zhao2015deep,lu2017deep,douze2016polysemous,zhao2015deep}.

\subsection{Comparison with the state of the art}
\label{stateart}
According to our results, one of the best performing visual descriptor is the ResNet and in particular SatResNet-50, while the best visual descriptor, when the computational cost is taken into account, is the Vgg M 128. We compared these descriptors, coupled with the four scheme described in Sec.~\ref{schamas}, with some recent methods~\cite{bosilj2016retrieval,aptoula2014,ozkan2014,yang2013}. All these works used the basic retrieval scheme and the experiments have been conducted on the LandUse dataset. \emph{Aptoula} proposed several global morphological texture descriptors~\cite{bosilj2016retrieval,aptoula2014}. \emph{Ozkan} et al. used bag of visual words (BoVW) descriptors, the vector of locally aggregated descriptors (VLAD) and the quantized VLAD (VLAD-PQ) descriptors~\cite{ozkan2014}. \emph{Yang} et al.~\cite{yang2013} investigated the effects of a number of design parameters on the BoVW representation. They considered: saliency-versus grid-based local feature extraction, the size of the visual codebook, the clustering algorithm used to create the codebook, and the dissimilarity measure used to compare the BOVW representations. 

The results of the comparison are shown in Table~\ref{tab:comparison_state_art}. The Bag of Dense SIFT (VLAD) presented in~\cite{ozkan2014}  achieves performance that is close to the CNN-based descriptors. This method  achieves $ANMRR= 0.460$ with $EQC= 5120$. This result has been obtained considering a codebook built by using images from the LandUse dataset. Concerning the computational cost, the texture features~\cite{yang2013,aptoula2014} are better than SatResNet-50 and Vgg M 128. In terms of trade-off between performance and computational cost, the Vgg M 128 descriptor achieves a $ANMRR$ value that is about 25\% lower than the one achieved by the  CCH+RIT+FPS$_{1}$+FPS$_{2}$ descriptor used in~\cite{aptoula2014} with a computational cost that is about 2 times higher.

\begin{table*}[tb]
\tiny
\centering
       \resizebox{0.9\textwidth}{!}{
  \centering
  \begin{tabularx}{\textwidth}{Zccccccccc}
    \toprule
 features 	&	Hist. Inters. & Euclidean&Cosine&Manhattan&$\chi^{2}$-square&Length&Time (sec)	&EQC\\
 \bottomrule
 CCH RIT FPS$_{1}$ FPS$_{2}$~\cite{aptoula2014}&0.609&0.640&-&0.589&0.575&62&-&12\\
 CCH~\cite{aptoula2014}&0.677&0.726&-&0.677&0.649&20&1.9&4\\
  RIT~\cite{aptoula2014}&0.751&0.769&-&0.751&0.757&20&2.3&4\\
 FPS$_{1}$~\cite{aptoula2014}&0.798&0.731&-&0.740&0.726&14&1.6&2\\
 FPS$_{2}$~\cite{aptoula2014}&0.853&0.805&-&0.790&0.783&8&1.6&1\\
pLPS-aug~\cite{ bosilj2016retrieval}&-&0.472&-&-&-&12288&-&2458\\
 \midrule
  Texture~\cite{yang2013}&-&0.630&-&-&-&-&40.4&-\\
    Local features~\cite{yang2013}&&0.591&-&-&-&-&193.3&-\\
     \midrule
  Dense SIFT (BoVW)~\cite{ozkan2014}&-&-&0.540&-&-&1024&9.4&204\\
  Dense SIFT (VLAD)~\cite{ozkan2014}&-&-&0.460&-&-&25600&129.3&5120\\
    \midrule
        B-IR Vgg M 128 &0.544&{0.488}&{0.488}&0.493&{0.488}&128&-&25\\
    B-IR ResNet-50 &0.476&0.358&0.358&0.395&{0.350}&2048&-&409\\
    B-IR SatResNet-50 &0.331&{0.239}&{0.239}&0.271&{0.233}&2048&-&409\\
        \midrule
    P-RF Vgg M 128 &0.550&0.470&0.470&0.466&{0.458}&128&-&125\\
    P-RF ResNet-50 &0.493&{0.305}&{0.305}&0.390&{0.324}&2048&-&2045\\
    P-RF SatResNet-50 &0.332&0.185&0.185&0.250&{0.200}&2048&-&2045\\
            \midrule
                M-RF Vgg M 128 &0.497&0.422&0.422&0.416&{0.410}&128&-&125\\
    M-RF ResNet-50  &0.459&0.181&0.181&0.359&{0.299}&2048&-&2045\\
    M-RF SatResNet-50 &0.307&\bf{0.014}&\bf{0.014}&\bf{0.224}&\bf{0.179}&2048&-&2045\\
         \midrule
               AL-RF Vgg M 128&{0.333}&-&-&-&-&128&-&500\\ 
 	 AL-RF ResNet-50&0.332&-&-&-&-&2048&-&8180\\
      AL-RF SatResNet-50&\bf{0.232}&-&-&-&-&2048&-&8180\\ 
           \bottomrule\\
  \end{tabularx}
  }
  \caption{ANMRR comparison on the LandUse dataset. The lower is the result, the better is the performance. For each column the best result is reported in bold.}
  \label{tab:comparison_state_art}
\end{table*}

\section{Conclusions}\label{conclusions}

In this work we presented an extensive  evaluation of  visual descriptors for content-based retrieval of remote sensing images. We evaluated global hand-crafted, local hand-crafted and Convolutional Neural Networks features coupled with four different content-based retrieval (CBIR) schemes: a basic CBIR, a pseudo relevance feedback (RF), a manual RF and an active-learning-based RF. The experimentation has been conducted on two publicly available datasets that are different in terms of image size and resolution. Results demonstrated that:
 \begin{itemize}
 \item CNN-based descriptors proved to perform better, on average, than both global hand-crafted and local hand-crafted descriptors whatever is the retrieval scheme adopted and on both the datasets considered, see the summary table~\ref{tab:summary};
 \item The RS domain adaptation of the ResNet-50 has led to a notable improvement of performance with respect to CNNs pre-trained on multimedia scene and object images. This demonstrated the importance of domain adaptation in the field of remote sensing images;
  \item NetVLAD works better on  those images that contain fine-grained textures and objects. NetVLAD is a CNN that considers local features. This is true specially for the LandUse dataset on classes like: chaparral, harbor, runaway, etc. See the tables~\ref{tab:class_land_alrf} and \ref{tab:class_rs_alrf} and figures~\ref{fig:bars};
 \item Pseudo and manual relevance feedback schemes demonstrated to be very effective only when coupled with a visual descriptor that is high performing in a basic retrieval system, such as CNN-based and local hand-crafted descriptors. This is quite evident looking at~\ref{fig:increment}a and b;
 \item Active-Learning-based RF demonstrated to be very effective on average and the best performing among retrieval schemes. The computational cost required to perform one query is, on average, at least 4 times higher than the computational cost required to perform a query with the other considered RF schemes and at least 20 times higher than a basic retrieval scheme.
 \end{itemize}

As future works, it would be interesting to experiments the efficiency of techniques to speed up the image search process by exploiting compact feature vector representations such as has, or polysemous codes.

\section*{Competing interests}

The author declare that he has no competing interests.

\section*{Funding}
We gratefully acknowledge the support of NVIDIA Corporation with the donation of the Tesla K40 GPU used for doing part of the experiments included in this research.

\section*{Acknowledgments}
The author is grateful to Prof. Raimondo Schettini for the valuable comments and stimulating discussions and  he would like to thank the reviewers for their valuable comments and effort to improve the manuscript. 

\bibliographystyle{tfcad}
\bibliography{GRSL-2014}   

\end{document}